\documentclass[10pt,twocolumn,letterpaper]{article}

\usepackage{cvpr}              

\definecolor{cvprblue}{rgb}{0.21,0.49,0.74}
\usepackage[pagebackref,breaklinks,colorlinks,allcolors=cvprblue]{hyperref}

\def\paperID{2022} 
\def\confName{CVPR}
\def\confYear{2025}

\usepackage{booktabs}
\usepackage{multirow}
\usepackage{algorithm}
\usepackage{algpseudocode}
\usepackage{xcolor}
\usepackage{tabularx}

\title{Finding Local Diffusion Schr\"odinger Bridge using Kolmogorov-Arnold Network
}

\author{
Xingyu Qiu$^{1}$, 
Mengying Yang$^{1}$, 
Xinghua Ma$^{1}$,
Fanding Li$^{1}$,
Dong Liang$^{1}$\\
\textbf{Gongning Luo$^{1*}$},	Wei Wang$^{2}$,	Kuanquan Wang$^{1}$, Shuo Li$^{3}$ \\
$^{1}$Harbin Institute of Technology, Harbin, China \\
$^{2}$Harbin Institute of Technology, Shenzhen, China \\
$^{3}$Case Western Reserve University, Cleveland, USA \\
\texttt{luogongning@hit.edu.cn}
}

\begin{document}
\maketitle
\begin{abstract}
	In image generation, Schrödinger Bridge (SB)-based methods theoretically enhance the efficiency and quality compared to the diffusion models by finding the least costly path between two distributions. However, they are computationally expensive and time-consuming when applied to complex image data. The reason is that they focus on fitting globally optimal paths in high-dimensional spaces, directly generating images as next step on the path using complex networks through self-supervised training, which typically results in a gap with the global optimum.
	Meanwhile, most diffusion models are in the same path subspace generated by weights $f_A(t)$ and $f_B(t)$, as they follow the paradigm ($x_t = f_A(t)x_{Img} + f_B(t)\epsilon$).
	To address the limitations of SB-based methods, this paper proposes for the first time to find local Diffusion Schrödinger Bridges (LDSB) in the diffusion path subspace, which strengthens the connection between the SB problem and diffusion models. 
	Specifically, our method optimizes the diffusion paths using Kolmogorov-Arnold Network (KAN), which has the advantage of resistance to forgetting and continuous output.
	The experiment shows that our LDSB significantly improves the quality and efficiency of image generation using the same pre-trained denoising network and the KAN for optimising is only \textbf{less than 0.1MB}. 
	The FID metric is reduced by \textbf{more than 15\%}, especially with a reduction of \textbf{48.50\%} when NFE of DDIM is $5$ for the CelebA dataset. Code is available at https://github.com/PerceptionComputingLab/LDSB.
\end{abstract}

\section{Introduction}
\begin{figure}[t]
	\centering
	\includegraphics[width=1\columnwidth]{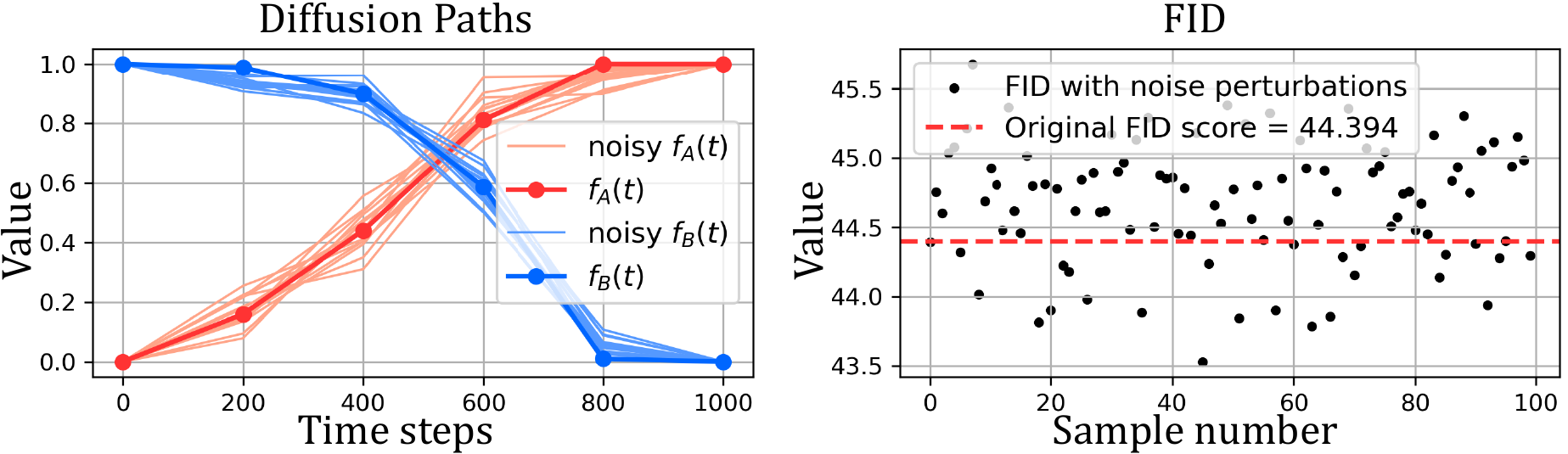} %
	\caption{Adjusting the weights of image and noise ($f_A(t)$ and $f_B(t)$) in diffusion enhances image generation quality, as the presence of perturbation reduces the FID. This paper achieves optimization by solving a local diffusion Schrödinger bridge in the diffusion path subspace generated by $f_A(t)$ and $f_B(t)$.}
	\label{moti}
\end{figure}
Optimizing the diffusion path is a key to improve the image generation quality which is usually ignored.
Recently, numerous studies have focused on generating higher quality images from Gaussian noise in fewer steps. 
They try to use different algorithmic frameworks to improve the generality and efficiency of diffusion models, such as Langevin dynamic sampling \cite{songscore,ho2020denoising,song2020denoising} and Continuous Normalizing Flows (CNFs) \cite{lipman2022flow,liu2022flow}. Specifically, they not only change the training objectives of the denoising network, but also the diffusion paths. 
In this paper, the \textbf{diffusion path} refers to the transfer trajectory containing multiple states between the prior distribution (Gaussian noise) and the target distribution (images). And the \textbf{denoising network} refers to the model that separates the noise from the image in the diffusion path.
The diffusion paths strongly influence the quality of image generation \cite{DSB,tang2024simplified}, however, they are predefined. For instance, DDPM \cite{ho2020denoising} and DDIM \cite{song2020denoising} use $\sqrt{\alpha_t} $ and $\sqrt{1-\alpha_t} $ as the weights for images $x$ and Gaussian noise $\epsilon$, respectively, SGM \cite{songscore} and NCSN \cite{song2019generative} use weights of $1$ and $\sigma_t$, and Flow Matching \cite{lipman2022flow} and Recited Flow\cite{liu2022flow} use $1-t$ and $t$, where $\alpha_t $ and $\sigma_t $ are predefined functions of $t$. In summary, their diffusion paths follow the paradigm ($x_t = f_A(t)x_{Img} + f_B(t)\epsilon$), and different methods assign different weights to the image $x_{Img}$ and Gaussian noise $\epsilon$, as illustrated in Tab. \ref{tab1}. 
To further investigate the impact of diffusion paths on image generation quality, a toy experiment introduces random noise into the weights of $x_{Img}$ and $\epsilon$. As shown in Fig. \ref{moti}, the presence of some perturbations does improve the quality of image generation. 
This paper aims to find the diffusion paths that improve the image generation quality based on Schrödinger Bridge, while keeping the denoising network unchanged.

The Schrödinger Bridge (SB)-based methods aim to find the optimal transport path between prior and target distributions, however, they encounter significant training challenges and struggle to approximate the globally optimal path in practice. The Schrödinger bridge \cite{schrodinger1932theorie,chetrite2021schrodinger} problem can theoretically identify the optimal transport between any two distributions, yielding more accurate samples in less time compared to diffusion models. Nonetheless, the Schrödinger problem rarely admits a closed-form solution \cite{DSB}. 
Diffusion Schrödinger Bridge (DSB) \cite{DSB} first considered generative model as a Schrödinger Bridge problem and solved it with Iterative Proportional Fitting (IPF) \cite{fortet1940resolution,kullback1968probability,ruschendorf1993note}.
However, finding the globally optimal path brings convergence challenges to DSB when dealing with complex data in high-dimensional space. Specifically, DSB iteratively trains two complex networks whose inputs and outputs are the target image size in an unsupervised way to transform between two distributions. In practice, there is still a gap between the DSB and the global optimal path. Subsequent works \cite{tang2024simplified,shi2024diffusion,gushchin2024light,huang2024one,huang2024schr} simplify the the diffusion Schrödinger bridge but still face challenges of time-consuming, expensive computational resources and complex path optimisation. 

In this paper, we propose to find local diffusion Schrödinger bridge (LDSB) within the path subspace $\mathcal{P}_{A,B}$ using Kolmogorov-Arnold Network, improving image generation quality and efficiency at a low computational cost. 
Specifically, LDSB models a diffusion path subspace which covers most diffusion models, generated by the weights of the image and noise, $f_A(t)$ and $f_B(t)$. Then LDSB iteratively optimises the $f_A(t)$ and $f_B(t)$ of both forward and reverse diffusion based on IPF to find the local diffusion Schrödinger bridge in subspace $\mathcal{P}_{A,B}$, achieving higher image generation quality. 
Unlike other Schrödinger bridge-based methods that train complex denoising networks, the neural network for optimising diffusion paths in our LDSB is only less than 0.1MB.
Training diffusion paths relies on previously learnt time steps and the effective diffusion paths are typically continuous. Thus, LDSB uses the Kolmogorov-Arnold Network (KAN) because of its resistance to forgetting and continuous output, which MLP lacks.
Our method was tested in two mainstream diffusion frameworks: the diffusion-based DDIM and the flow-based Recited flow. And our experiments focused on improving image generation quality with fewer sampling steps (5, 10 and 20 step). Qualitative and quantitative comparisons show that LDSB significantly enhances the quality and efficiency of image generation with the same pre-trained denoising networks.

In summary, our main contributions are as follows.
\begin{itemize}
	\item For the first time, we optimise the weights of image and noise in the diffusion paths based on the Schrödinger bridge and propose the Local Diffusion Schrödinger Bridge (LDSB), which simplifies the training and enhances the image generation.
	\item LDSB models a diffusion path subspace reparametrically to achieve more accurate connections between time steps, and proposes the first optimisation strategy for the weights in diffusion paths based on IPF.
	\item Quantitative and qualitative experiments validate that LDSB improves the image generation quality and efficiency using the same pre-trained denoising networks.
\end{itemize}
\section{Related Work}

\subsection{Most Diffusion Paths are Manually Predetermined But not Optimisable}
To improve the image generation, researches focus on proposing various algorithmic frameworks, but neglecting the importance of diffusion paths. Diffusion paths are always pre-determined manually but not optimised. 
The Score-Based Generative Modeling (SGM) \cite{songscore} introduces different degrees of Gaussian noise to create diffusion path between the images $x_0$ and Gaussian noise $\epsilon$ based on Langevin dynamic sampling across diffusion steps $t \in \{0,\dots,T\}$. And It gradually generates images from noise along the diffusion path. The degree of Gaussian noise added is artificially set, with two examples provided: Variance Exploding (VE) and Variance Preserving (VP). The VP is:
\begin{equation}
	\begin{split}
		&x_t=\sqrt{\overline{\alpha}_t} x_0+ \sqrt{1 - \overline{\alpha}_t} \epsilon\\
		&x_{t+1} = \sqrt{1-\beta_{t+1}}x_t+\sqrt{\beta_{t+1}}\epsilon		
	\end{split}
	\label{ddim}
\end{equation}
where $\overline{\alpha}_t = \prod_{i=1}^t (1-\beta_i)$, and $\beta_t$ is an artificially set function of $t$, which is linear, constant, or sigmoid, ensuring that $\sqrt{\overline{\alpha}_0} \approx 1$ and $\sqrt{\overline{\alpha}_T} \approx 0$. It is similar to DDPM\cite{ho2020denoising} and DDIM\cite{song2020denoising}.

The VE is:
\begin{equation}
	\begin{split}
		&x_t = x_0+\sigma_t \epsilon\\
		&x_{t+1}=x_t+\sqrt{\sigma_{t+1}^2-\sigma_t^2}\epsilon
	\end{split}
\end{equation}
the larger $\sigma$ is better, as $x_T$ should be as close to the noise as possible. It is similar to NCSN \cite{song2019generative}.

Flow matching \cite{song2019generative} enhances the generality of the diffusion model based on the Continuous Normalizing Flows (CNFs). It overcomes the limitations of the necessity to use Gaussian noise and theoretically achieves the transformation of any image domain. For formal consistency, images are still generated from Gaussian noise in the image generation task. It emphasizes a natural and intuitive linear diffusion path in original paper:
\begin{equation}
	x_t = (1-t) x_0+(t+\sigma_{min}(1-t))\epsilon
	\label{FM}
\end{equation}
In subsequent work, Recited Flow \cite{liu2022flow} considers the $\sigma_{min}$ is $0$, focusing on the potential of linear paths to improve sampling efficiency.

In addition, there are some other studies \cite{bansal2024cold,hoogeboom2022blurring,liu2024residual,liu2023i2sb,chen2023schrodinger,karras2022elucidating,nichol2021improved,luo2023image} also utilising different diffusion paths. In summary, the diffusion paths can be modeled as a general paradigm: 
\begin{equation}
	\label{path paradigm}
	x_t=f_A(t)x_0+f_B(t)\epsilon
\end{equation}
different methods define the diffusion path with distinct $f_A(t)$ and $f_B(t)$, placing them within the diffusion path subspace $\mathcal{P}_{f_A,f_B} $ generated by $f_A$ and $f_B$.

\begin{table}[htpb]
	\centering
	\caption{Several classical diffusion paths in the subspace $\mathcal{P}_{f_A,f_B}$ follow the paradigm $x_t=f_A(t)x_0+f_B(t)\epsilon$.}
	\begin{tabular}{c|cc} 
		\toprule
		\multicolumn{1}{c}{Diffusion Path} & \multicolumn{1}{c}{$f_A(t)$} & \multicolumn{1}{c}{$f_B(t)$}\\
		\midrule													
		Variance Exploding (VE) \cite{songscore,song2019generative} & $1$ &$\sigma_t$ \\
		Variance Preserving (VP) \cite{songscore,ho2020denoising,song2020denoising} & $\sqrt{\overline{\alpha}_t}$ & $ \sqrt{1 - \overline{\alpha}_t}$ \\
		Linear Path \cite{lipman2022flow,liu2022flow}& $1-t$ & $t$ \\
		\bottomrule
	\end{tabular}
	\label{tab1}
\end{table}

\subsection{Schrödinger Bridge Auto Find Diffusion Paths But face Challenges}
The Schrödinger Bridge (SB) problem \cite{schrodinger1932theorie,chetrite2021schrodinger} aims to find the least costly path between two distributions, but in practice it suffers from training difficulties and struggles to improve performance on complex data. Specifically, it is to find the optimal solution $\pi^*$ of the following optimization problem:
\begin{equation}
	\begin{split}
		\pi^*=arg min\{& KL(\pi|p_{ref}):\pi \in \mathcal{P}_{N+1}\\
		,& \pi_0=P_{data},\pi_N=p_{prior}\} \\\
	\end{split}
\end{equation}
where $\mathcal{P}_{N+1}$ represents the path space, $\pi_0$ and $\pi_N$ denote the distributions at both ends of the path, and $p_{ref}$ is the preset reference path. If $\pi^*$ is available, it can be sampled from $p_{prior}$ to $p_{data}$ or from $p_{data}$ to $p_{prior}$.

Typically, Schrödinger Bridge problem lacks closed-form solution. Most works use Iterative Proportional Fitting (IPF) \cite{fortet1940resolution,kullback1968probability,ruschendorf1993note} to approximate the Schrödinger bridge, which is optimised iteratively:
\begin{equation}
	\begin{split}
		&\pi^{2n+1}=arg min\{KL(\pi|\pi^{2n}):\pi \in \mathcal{P}_{N+1},\pi_N=p_{prior}\}\\
		&\pi^{2n+2}=arg min\{KL(\pi|\pi^{2n+1}):\pi \in \mathcal{P}_{N+1},\pi_0=P_{data}\}\\
	\end{split}
	\label{IPF}
\end{equation}
\textbf{Intuition: }The end of $\pi^{2n+1}$ is $p_{prior}$, and the start of $\pi^{2n+2}$ is $p_{data}$. 
Through iterations, $\pi^{2n+1}$ and $\pi^{2n+2}$ become similar increasingly, with the final paths beginning at $p_{prior}$ and ending at $p_{data}$.\\
The initial $\pi^0$ is a preset reference path. However, IPF requires calculating the joint probability density to obtain $\pi^n$, which significantly increases computational complexity and makes it impractical \cite{DSB,tang2024simplified}.

The Diffusion Schrödinger Bridge (DSB) \cite{DSB} is an approximate implementation of IPF. It decomposes the joint density into a series of conditional density optimization problems. Specifically, the path $\pi$ is decomposed into $\pi_{k+1|k}$ and $\pi_{k|k+1}$ by introducing a time step $t$ in the diffusion model:
\begin{equation}
	\begin{split}
		\pi^{2n+1}=arg min\{&KL(\pi_{k|k+1}|\pi^{2n}_{k|k+1}):\\
		&\pi \in \mathcal{P}_{N+1}, \pi_N=p_{prior}\}\\
		\pi^{2n+2}=arg min\{&KL(\pi_{k+1|k}|\pi^{2n+1}_{k+1|k}):\\
		&\pi \in \mathcal{P}_{N+1},\pi_0=P_{data}\}\\
	\end{split}\label{DIPF}
\end{equation}
In practice, DSB employs two independent neural networks, $F^n_k$ and $B^n_k$, to predict each transformation step of the forward and reverse diffusion, respectively. Through mathematical derivations and approximations \cite{DSB}, the training loss of DSB is formulated as follows: 
\begin{equation}
	\begin{split}
		L_{B^n_{k+1}}&=\mathbb{E}_{(x_k,x_{k+1})\sim p^n_{k,k+1}}\\
		&\left[||B^n_{k+1}(x_{k+1})-(x_{k+1}+F^n_k(x_k)-F^n_k(x_{k+1}))||^2\right]\\
		L_{F^{n+1}_{k}}&=\mathbb{E}_{(x_k,x_{k+1})\sim q^n_{k,k+1}}\\
		&\left[||F^{n+1}_{k}(x_{k})-(x_{k}+B^n_{k+1}(x_{k+1})-B^n_{k+1}(x_{k}))||^2\right]\\
	\end{split}
\end{equation}
where $p^n$ and $q^n$ represent the forward and reverse joint probability distributions, respectively, corresponding to $\pi_{2n}$ and $\pi_{2n+1}$ in the Eq. \ref{DIPF}. The DSB iteratively optimizes the $F^n_k$ and $B^n_k$ for $n \in {0, 1, …, L}$ to minimize Eq.\ref{DIPF}.

Training DSB is challenging with complex data. The network of the DSB is trained by the inference of another network and the unsupervised search for the diffusion path. 
In addition, the large input and output sizes broaden the path search range $\mathcal{P}_{N+1}$ to global levels, increasing network complexity and resulting in challenging, time-consuming, and computationally costly optimization \cite{tang2024simplified}.

Subsequently, the Simplified Diffusion Schr\"odinger Bridge (S-DSB) \cite{tang2024simplified} is dedicated to simplifying and accelerating the diffusion Schrödinger bridge. It simplifies the training loss and reduces the amount of computation, as:
\begin{equation}
	\begin{split}
		L_{B^n_{k+1}}=&\mathbb{E}_{(x_k,x_{k+1})\sim p^n_{k,k+1}}\left[||B^n_{k+1}(x_{k+1})-x_{k}||^2\right]\\
		L_{F^{n+1}_{k}}=&\mathbb{E}_{(x_k,x_{k+1})\sim q^n_{k,k+1}}\left[||F^{n+1}_{k}(x_{k})-x_{k+1}||^2\right]\\
	\end{split}
\end{equation}
However, S-DSB still relies on unsupervised and iterative training of two complex networks, and faces similar challenges with training difficulty as DSB. 
This paper strengthens the connection between Schr\"odinger Bridge and diffusion models by solving the SB problem within path subspace $\mathcal{P}_{f_A,f_B}$ (optimising the $f_A$ and $f_B$ in Eq. \ref{path paradigm}), while keeping the pre-trained denoising network unchanged.

\subsection{KAN is Capable of Predicting Diffusion Path}
The Kolmogorov-Arnold Network (KAN) excels in predicting diffusion paths. According to the Kolmogorov-Arnold theorem \cite{kolmogorov1957representation}, any continuous function can be represented as a combination of continuous unary functions of finite variables. Leveraging this theorem, KAN \cite{liu2024kan} is introduced as an innovative network architecture. Unlike the Multilayer Perceptron (MLP), which optimizes a static weight matrix and employs a fixed nonlinear activation function, KAN utilizes multiple spline functions as its basis and optimizes the univariate weights of these nonlinear basis functions as learnable activation functions. 

KAN is particularly suitable for optimizing diffusion paths because of (1) resistance to forgetting and (2) its output is continuous.
(1) The local plasticity of KAN helps to prevent catastrophic forgetting, since the spline basis function is local, a sample only influences nearby spline function weights, leaving distant spline weights unaffected. In contrast, MLPs typically use global activations, where local changes propagate uncontrollably to distant areas, disrupting stored information. 
Our method employs a step-by-step optimization method similar to the Diffusion Schrödinger Bridge (DSB) during training, such that the current path step $t$ is optimized based on the previous steps $[0,\dots, t-1]$. The resistant forgetting capability of the KAN ensures that the previous steps is remained as possible when optimising the current step. 
(2) The continuous output of KAN arises from the inherent continuity of spline functions. An efficient diffusion path requires a smooth transition from the prior to the target domain. The continuous output of KAN aligns closely with smooth transition, making it more consistent with the diffusion path.

\section{Method}
To find local diffusion Schrödinger bridges (LDSB), Sec. \ref{3.1} reparameterizes the diffusion path and defines the diffusion path subspace $\mathcal{P}_{A,B}$, Sec. \ref{3.2} introduces training objectives to fit local Schrödinger bridges within the subspace $\mathcal{P}_{A,B}$, and Sec. \ref{3.3} determines the initial states of the forward and reverse diffusion paths.
\subsection{Reparameterized Modeling of Diffusion Path}
\label{3.1}
In order to capture the relationship between each time step $t$ in the diffusion process more accurately, the diffusion paths are reparametrically modelled.
Eq. \ref{path paradigm} summarizes the commonly used forms of diffusion paths in current methods. Different diffusion paths correspond to different weights $f_A$ and $f_B$, yet they share common features. For instance in forward diffusion, $f_A(t)$ is monotonically decreasing, and $f_B$ is monotonically increasing, ensuring that Eq. \ref{path paradigm} gradually transforms from image to noise and vice versa. Consequently, our method reparameterizes $f_A(t)$ as $\prod_{i=0}^t A(i)$, and $f_B(t)$ as $\prod_{i=T}^t B(i)$, where $A(i)$ and $B(i)$ are usually less than $1$. The $\mathcal{P}_{A,B}$ is defined as the diffusion path subspace generated by $A$ and $B$:
\begin{equation}
	\begin{split}
		\mathcal{P}_{A,B} = \{\{x_t\}_{t=0}^T \mid x_t = \prod_{i=0}^t A(i) x_0 + \prod_{i=T}^t B(i) \epsilon,\\
		\forall t,i \in \{0, \dots, T\}, A(i), B(i) \in \mathbb{R}\}\\
	\end{split}
\end{equation}

The $\pi^{2n+1}$ and $\pi^{2n+2}$ represents the reverse and forward diffusion path, respectively, which is consistent with the DSB Eq. \ref{DIPF}:
\begin{equation}
	\begin{split}
		&\pi^{2n+1}: x_t^{2n+1} = \prod_{i=0}^t A^{2n+1}(i) x_{0_\theta} + \prod_{i=T}^t B^{2n+1}(i) \epsilon_\theta\\
		&\pi^{2n+2}: x_t^{2n+2} = \prod_{i=0}^t A^{2n+2}(i) x_{0} + \prod_{i=T}^t B^{2n+2}(i) \epsilon\\
		&\forall n: \pi^{2n+1},\pi^{2n+2} \in \mathcal{P}_{A,B}
	\end{split}
\end{equation}
where the $n$ is the number of iterative optimisation.
For conciseness, the $x_t$ of both forward and reverse paths are represented by the image $x_0$ and the noise $\epsilon$. The $x_{0_\theta}$ and $\epsilon_\theta$ of the reverse path are predicted from the $P{prior}$ using the network, see appendix for details. 

Reparameterizing the diffusion path has two advantages compared to using Eq. \ref{path paradigm} directly. 
First, the correlations between each time states $x_t$ are modeled more exactly. 
The adjacent time steps of the reparameterized path, such as $\prod_{i=0}^t A(i)$ and $\prod_{i=0}^{t+1} A(i)$, are multiplied by $A(t+1)$, whereas they are independent of each other in Eq. \ref{path paradigm}.
Second, the optimization is more robust. 
For instance, the prediction of $f_B(t)$, which is monotonic, benefits from knowing $f_B(t+1)$. The reparameterized path $B(t)$ has a prediction range of $(0, 1)$ , while Eq. \ref{path paradigm} has a much narrower prediction range of $(f_B(t+1), 1)$. Given the same prediction error, the reparameterized path is more robust.

\subsection{Training Target and Path Analysis}
\label{3.2}
The IPF and DSB move the paths $\pi_{2n+1}$ and $\pi_{2n+2}$ close enough in Eq. \ref{IPF} and \ref{DIPF}. However, they are difficult and time consuming to train as optimizing the complex data in global space as next time state with unsupervised iterations.

To address this problem, our method proposes to find local Schrödinger bridges within a smaller diffusion paths subspace $\mathcal{P}_{A,B}$, as defined in Sec. \ref{3.1}. 
By optimizing only the weights $A$ and $B$ in diffusion paths while keeping the pre-trained denoising network unchanged, our method enhances the efficiency and quality of image generation with fewer computations and less training complexity.
Specifically, the reverse $\pi^{2n+1}$ and forward $\pi^{2n+2}$ paths are iteratively optimized to make the $x_t^{2n+1}$ and $x_t^{2n+2}$ as similar as possible for each time step $t$. The training objectives for $\pi^{2n+1}$ and $\pi^{2n+2}$ are presented in the following.
\begin{figure}[htbp]
	\centering
	\includegraphics[width=\columnwidth]{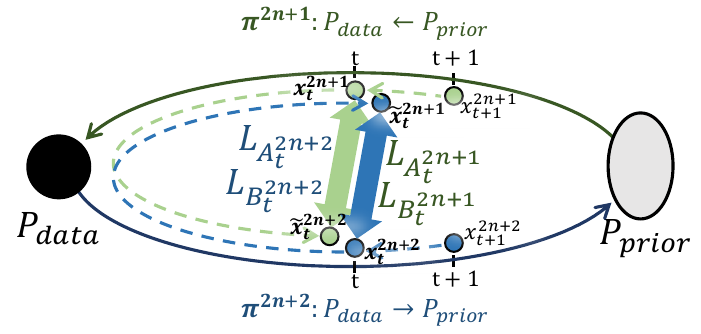} %
	\caption{Visualisation of training objectives. \textcolor{blue}{Blue dotted lines and arrow} represent the optimization of the inverse path $\pi^{2n+1}$ and \textcolor{ForestGreen}{green dotted lines and arrow} for the forward path $\pi^{2n+2}$.}
	\label{Outputs}
\end{figure}
\subsubsection{Optimizing $\pi^{2n+1}$ using $\pi^{2n}$}
Given the $\{x_t^{2n}\}_{t=0}^T$ on $\pi^{2n}$, LDSB first goes from $x_{t+1}^{2n}$ to $x_t^{2n+1}$ along the diffusion path:
\begin{equation}
	\begin{split}
		\tilde{x}_{t}^{2n+1}=& \frac{\prod_{i=0}^t A^{2n+1}(i)}{\prod_{i=0}^{t+1} A^{2n}(i)}\left(x_{t+1}^{2n}-\prod_{i=T}^{t+1} B^{2n}(i) \epsilon_\theta\right)\\
		&+ \prod_{i=T}^t B^{2n+1}(i) \epsilon_\theta\\
	\end{split}
	\label{node}
\end{equation}
where the $\epsilon_\theta$ is the predicted noise predicted by the pre-trained model $\theta$ with input $x_{t+1}^{2n}$ and $t+1$, as $\epsilon_\theta(x_{t+1}^{2n},t+1)$. And the $x_{t+1}^{2n}$ is generated by the sampled Gaussian noise $\epsilon'$ and images $x_0$ along  $\pi^{2n}$.
The reason for using $\epsilon_\theta$ to optimise the reverse path $\pi^{2n+1}$ is that the pre-trained model $\theta$ is an essential part of the reverse sampling process.

\textbf{Intuition of Eq. \ref{node}: }$x_{t+1}^{2n}$ lies on the forward path $\pi^{2n}$, while $x_t^{2n+1}$ lies on the reverse path $\pi^{2n+1}$. Moving from $x_{t+1}^{2n}$ to $x_t^{2n+1}$ requires a transit at $x_0$, which is the intersection of the two paths. Specifically, first follow $\pi^{2n}$ from $x_{t+1}^{2n}$ to $x_0$ (where $\frac{x_{t+1}^{2n}-\prod_{i=T}^{t+1} B^{2n}(i)\epsilon_\theta}{{\prod_{i=0}^{t+1} A^{2n}(i)}}$ represents $\tilde{x}_0$), and then along $\pi^{2n+1}$ from $x_0$ to $x_t^{2n+1}$ ($\prod_{i=0}^t A^{2n+1}(i) \tilde{x}_0 +\prod_{i=T}^t B^{2n+1}(i) \epsilon_\theta$), as shown by the \textcolor{blue}{blue dotted line} in Fig. \ref{Outputs}.

To optimise $\pi^{2n+1}$ to be similar to $\pi^{2n}$, the distance between $x_{t}^{2n}$ and $\tilde{x}_{t}^{2n+1}$ is minimized at each time step $t$:
\begin{equation}
	L_{\pi^{2n+1}}=\sum_{t=1}^{T-1}||\tilde{x}_{t}^{2n+1}(A^{2n+1},B^{2n+1})-x_t^{2n}||^2
\end{equation}

After reparameterisation, the $A^{2n+1}_t$ and $B^{2n+1}_t$ are optimized by $L_{A^{2n+1}_t}$ and $L_{B^{2n+1}_t}$ respectively:
\begin{equation}
	\begin{split}
		&L_{B^{2n+1}_t}=\sum_{t=1}^{T-1}||B^{2n+1}(t)-\tilde{B}^{2n+1}_t||^2\\
		&\tilde{B}^{2n+1}_t=\frac{x_t^{2n}-\frac{\prod_{i=0}^t A^{2n+1}(i)}{\prod_{i=0}^{t+1} A^{2n}(i)}\left(x_{t+1}^{2n}-\prod_{i=T}^{t+1} B^{2n}(i)\epsilon_\theta\right)}{\prod_{i=T}^{t+1} B^{2n+1}(i) \epsilon_\theta}\\
	\end{split}
	\label{LB}
\end{equation}
\begin{equation}
	\begin{split}
		&L_{A^{2n+1}_t}=\sum_{t=1}^{T-1}||A^{2n+1}(t)-\tilde{A}^{2n+1}_t||^2\\
		&\tilde{A}^{2n+1}_t=\frac{x_t^{2n}-\prod_{i=T}^t B^{2n+1}(i) \epsilon_\theta}{\frac{\prod_{i=0}^{t-1} A^{2n+1}(i)}{\prod_{i=0}^{t+1} A^{2n}(i)}\left(x_{t+1}^{2n}-\prod_{i=T}^{t+1} B^{2n}(i)\epsilon_\theta\right)}\\
	\end{split}
	\label{LA}
\end{equation}

\subsubsection{Optimizing $\pi^{2n+2}$ using $\pi^{2n+1}$}
To optimise $\pi^{2n+2}$ to be similar to $\pi^{2n+1}$, the distance between $x_{t}^{2n+1}$ and $\tilde{x}_{t}^{2n+2}$ is minimized at each time step $t$.
Then similarly, it needs to go from $x_{t+1}^{2n+1}$ to $x_{t}^{2n+2}$ first:
\begin{equation}
	\begin{split}
		\tilde{x}_{t}^{2n+2}= &\frac{\prod_{i=0}^{t} A^{2n+2}(i)}{\prod_{i=0}^{t+1} A^{2n+1}(i)}\left(x_{t+1}^{2n+1}-\prod_{i=T}^{t+1} B^{2n+1}(i) \epsilon'\right)\\
		&+\prod_{i=T}^{t} B^{2n+2}(i) \epsilon'\\
	\end{split}
	\label{node1}
\end{equation}
The $\epsilon'$ is the sampled Gaussian noise. And $x_{t+1}^{2n+1}$ is generated from $\epsilon'$ along the $\pi^{2n+1}$ using denoisng network $\theta$.

\textbf{Intuition of Eq. \ref{node1}: }Moving from $x_{t+1}^{2n+1}$ to $x_{t}^{2n+2}$ requires a transit at $x_0$, which is the intersection of the $\pi^{2n+1}$ and $\pi^{2n+2}$, as shown by the \textcolor{ForestGreen}{green dotted line} in Fig. \ref{Outputs}.

The training objective of optimizing $\pi^{2n+2}$ is:
\begin{equation}
	L_{\pi^{2n+2}}=\sum_{t=1}^{T-1}||\tilde{x}_{t}^{2n+2}(A^{2n+2},B^{2n+2})-x_{t}^{2n+1}||^2
\end{equation}

After reparameterisation, the $L_{B^{2n+2}_{t}}$ and $L_{A^{2n+2}_{t}}$ are:
\begin{equation}
	\begin{split}
		&L_{B^{2n+2}_{t}}=\sum_{t=1}^{T-1}||B^{2n+2}(t)-\tilde{B}^{2n+2}_{t}||^2\\
		&\tilde{B}^{2n+2}_{t}=\frac{x_{t}^{2n+1}-\frac{\prod_{i=0}^{t} A^{2n+2}(i)}{\prod_{i=0}^{t+1} A^{2n+1}(i)}\left(x_{t+1}^{2n+1}-\prod_{i=T}^{t+1} B^{2n+1}(i)\epsilon'\right)}{\prod_{i=T}^{t+1} B^{2n+2}(i) \epsilon'}\\
	\end{split}
	\label{LB1}
\end{equation}

\begin{equation}
	\begin{split}
		&L_{A^{2n+1}_{t}}=\sum_{t=1}^{T-1}||A^{2n+2}(t)-\tilde{A}^{2n+2}_{t}||^2\\
		&\tilde{A}^{2n+2}_{t}=\frac{x_{t}^{2n+1}-\prod_{i=T}^{t} B^{2n+2}(i) \epsilon'}{\frac{\prod_{i=0}^{t-1} A^{2n+2}(i)}{\prod_{i=0}^{t+1} A^{2n+1}(i)}\left(x_{t+1}^{2n+1}-\prod_{i=T}^{t+1} B^{2n+1}(i)\epsilon'\right)}\\
	\end{split}
	\label{LA1}
\end{equation}

To summarize, the training objectives are proposed to find local Schrödinger bridges within the diffusion path subspace $\mathcal{P}_{A,B}$. The optimization of diffusion paths by LDSB is described in Algorithm \ref{al1}.

\subsection{Diffusion Path Initialization and Optimization}
\label{3.3}
The forward and reverse paths in the LDSB are initialised with the original paths.
The training data of the pre-trained denoising network is generated from the original paths $\pi^0$. For instance, DDIM is generated by Eq. \ref{ddim} and FM by Eq. \ref{FM}. 
Our method aims to improve image generation quality by keeping denoising model unchanged and only optimizing the diffusion paths.
It allows the optimized path $\pi^{2L+1}$ remains close to the original path $\pi^0$, otherwise the data generated from $\pi^{2L+1}$ would differ significantly from the training data.
Consequently, this method intuitively selects the original path as the initial path and optimizes the diffusion path based on the pre-trained denoising network.

In the proposed diffusion path subspace $\mathcal{P}_{A,B}$, the LDSB optimises the diffusion path based on the error of the denoising network. Proof by contradiction, if the network is sufficiently accurate, such as $\epsilon_\theta=\epsilon'$, Eq. \ref{node} and Eq. \ref{node1} will both equal zero, and the initial path will be the optimal path given the pre-trained denoising network.
Errors of the denoising network also lead to discrepancies between the forward and reverse diffusion paths.
Thus our LDSB can also be seen as optimizing the diffusion paths \textbf{based on the error of pre-trained denoising network} to improve image generation efficiency and quality.

\begin{algorithm}[htbp]
	
	\caption{Local Diffusion Schr\"odinger Bridge (LDSB)}
	\begin{algorithmic}[1]		
		\State Initialize diffusion path as $\alpha^{0}, \beta^{0}$
		
		\For {$n \in \{0,\dots,L\}$}
		\While{not converged} \Comment{Optimize the $\pi^{2n+1}$}
		\State Sample $\{x_{t}^{2n}\}_{t=0}^{T}$ along $\pi^{2n}$, where $X_T^{2n} \sim P_{prior}$
		\State Compute $\{\tilde{x}_{t}^{2n+1}\}_{t=1}^{T-1}$ using Eq. \ref{node}
		\State Compute $L_{B_t^{2n+1}}$ approximating Eq. \ref{LB} 
		\State $\beta^{2n+1} \leftarrow$ Gradient Step $L_{B_t^{2n+1}}$
		\State Compute $L_{A_t^{2n+1}}$ approximating Eq. \ref{LA} 
		\State $\alpha^{2n+1} \leftarrow$ Gradient Step $L_{A_t^{2n+1}}$
		\EndWhile
		\While{not converged} \Comment{Optimize the $\pi^{2n+2}$}
		\State Sample $\{x_{t}^{2n+1}\}_{t=0}^{T}$ along $\pi^{2n+1}$, where $X_0^{2n+1} \sim P_{data}$
		\State Compute $\{\tilde{x}_{t}^{2n+2}\}_{t=1}^{T-1}$ using Eq. \ref{node1}
		\State Compute $L_{B_t^{2n+2}}$ approximating Eq. \ref{LB1} 
		\State $\beta^{2n+2} \leftarrow$ Gradient Step $L_{B_t^{2n+2}}$
		\State Compute $L_{A_t^{2n+2}}$ approximating Eq. \ref{LA1} 
		\State $\alpha^{2n+2} \leftarrow$ Gradient Step $L_{A_t^{2n+2}}$
		\EndWhile
		
		\EndFor
		
		\State \Return $(\beta^{2L+1},\alpha^{2L+1})$

	\end{algorithmic}	\label{al1}
\end{algorithm}

\section{Experiments}
The experiments focus on efficient sampling with a lower number of function evaluations (NFE). Our LDSB method keeps the pre-trained denoising network unchanged and only optimizes the diffusion paths using KAN, which significantly improves image generation efficiency and quality at a low computational cost. Specifically, the size of the KAN used to predict the diffusion paths is less than 0.1MB.

\subsection{Baselines and Experimental Setup}
The experiments demonstrate the enhancement of LDSB for generation performance on two mainstream efficient generation frameworks: the diffusion-based DDIM \cite{song2020denoising} and the flow matching \cite{lipman2022flow} based on flow models. 
A toy experiment with NFE of $20$ is performed on a 2D checkerboard as shown in Fig. \ref{checkerboard}.
Three datasets of different resolutions are used for comparison, including CIFAR10 \cite{krizhevsky2009learning} ($32\times32$), CelebA \cite{liu2015deep} ($64\times64$) and CelebA-HQ \cite{karras2017progressive} ($256\times256$). 
To ensure experimental rigor and reproducibility, experiments employ publicly available pre-trained denoising networks which are officially provided in DDIM \cite{song2020denoising} and FM \cite{lipman2022flow}, and only changed the weights in diffusion paths. 
The experiments use pre-trained models of DDIM in CIFAR10 and CelebA, and FM in CIFAR10 and CelebA-HQ, while CelebA-HQ for DDIM and CelebA for FM are not provided officially.
The number of iterations $L$ is set to $3$.
The quality of the generated images is measured using the Fréchet Inception Distance (FID) \cite{heusel2017gans} metric, with $50,000$ generated images compared to all real images in dataset. The lower FID is better.

\begin{table}[htpb]
	\centering
	\caption{Comparison of LDSB optimised paths using MLP and KAN on the CIFAR10 dataset. KAN outperforms MLP in optimising diffusion paths. }
	\begin{tabular}{c|ccc} 
		\toprule
		\multicolumn{1}{c}{NFE}&5&10&20\\
		\midrule	
		\multicolumn{1}{c}{Original} & 44.394 & 18.678 & 11.080\\			
		\multicolumn{1}{c}{MLP}& 66.884&21.854&15.328\\
		\multicolumn{1}{c}{KAN}&\textbf{26.929}& \textbf{9.619}&  \textbf{7.073}\\
		\bottomrule
	\end{tabular}
	\label{MLP}
\end{table}

Our method employs KAN to predict diffusion paths. Because it exhibits resistance to forgetting due to the local plasticity of the spline function, which MLP does not have. 
This property enables KAN to outperform MLP in optimizing diffusion paths, as shown in Tab. \ref{MLP}. Our experiment sets the grid size of the spline function to double the total steps of the path, resulting in a more distinct spline function. For example, the grid size for predicting a 5-step path is set to $10$. Furthermore, compared to other Schrödinger bridge-based methods that train complex networks to output the next image, our trained KAN model is under 0.1 MB as shown in Tab. \ref{size}, significantly reducing training difficulty and computational resource consumption. Specifically, the nodes number in each layer of the KAN is set to [1, 4, 16 ,4 ,1].
\begin{table}[htpb]
	\centering
	\caption{The sizes of the KANs used to optimise paths with different total steps are all less than 0.1 MB.}
	\begin{tabular}{c|ccc} 
		\toprule
		\multicolumn{1}{c}{Total Steps}&5&10&20\\
		\midrule													
		\multicolumn{1}{c}{KAN Size}&0.040 MB&0.057 MB&0.091 MB\\
		\bottomrule
	\end{tabular}
	\label{size}
\end{table}

\subsection{Experimental Results}
To verify the performance improvement for efficient sampling with fewer NFE, the experiments report the enhancement of image generation with NFE of $5$, $10$, and $
20$. LDSB optimizes only the weights $f_A$ and $f_B$, representing diffusion paths, while using the official pre-trained model in DDIM \cite{song2020denoising} and Flow matching \cite{lipman2022flow}. 
Experiments demonstrate that LDSB achieves better image generation quality with the same sampling steps and accelerates the image generation efficiency of the diffusion model. 
\begin{figure}[htbp]
	\centering
	\includegraphics[width=\columnwidth]{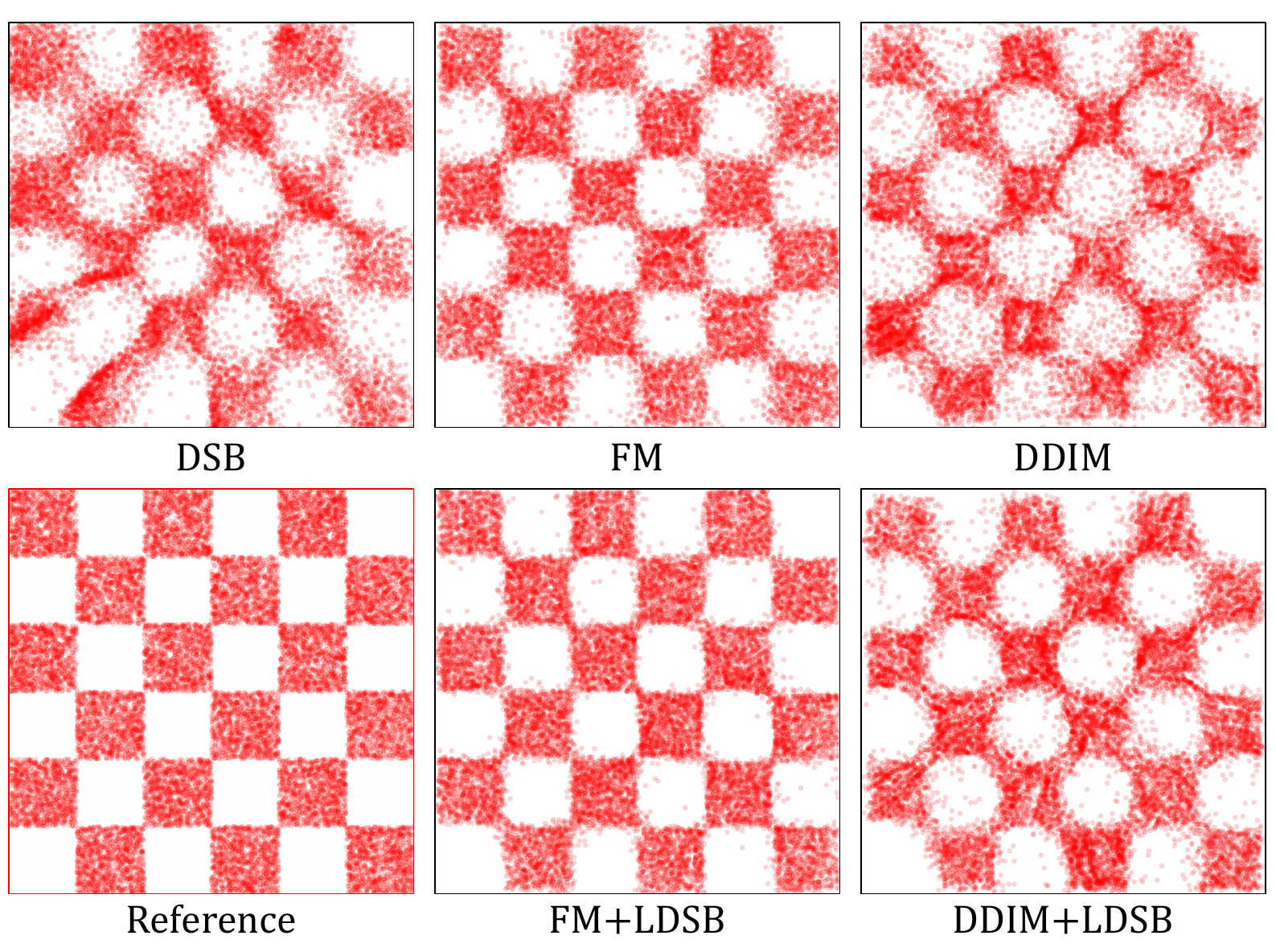}
	\caption{Comparison of different methods on 2D checkerboard with NFE of $20$. Optimisation with LDSB generates the checkerboard more accurately and with less error points. And LDSB in combination with DDIM or FM outperforms DSB. }
	\label{checkerboard}
\end{figure}
\begin{figure}[htbp]
	\centering
	\includegraphics[width=\columnwidth]{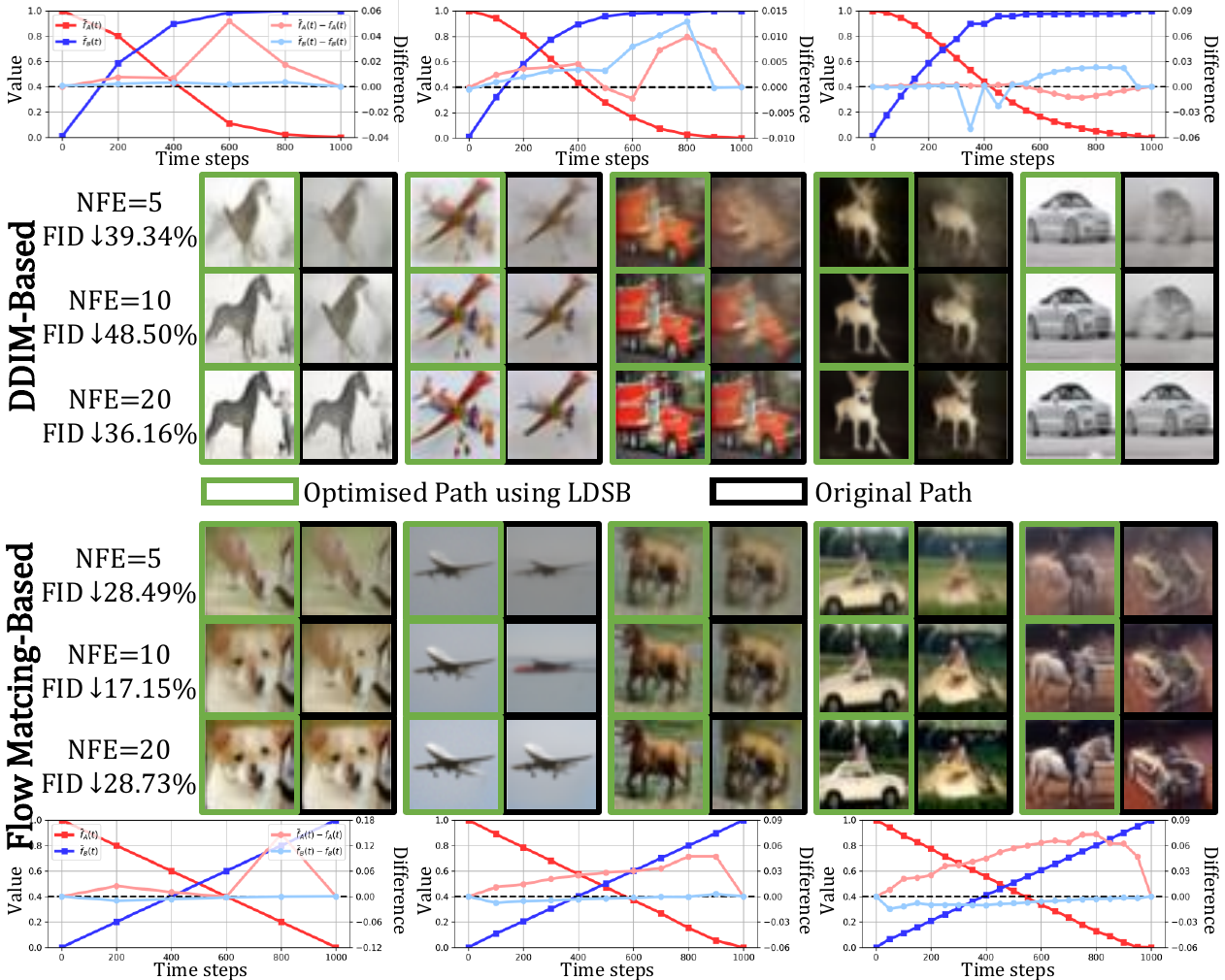}
	\caption{Comparison of diffusion paths of LDSB and the generated image on CIFAR10. LDSB makes the images more realistic and clearer on same NFE. $f(t)$ is the original diffusion path, and $\tilde{f}(t)$ is the optimised path using LDSB.}
	\label{Cifar10}
\end{figure}
\begin{figure}[htbp]
	\centering
	\includegraphics[width=\columnwidth]{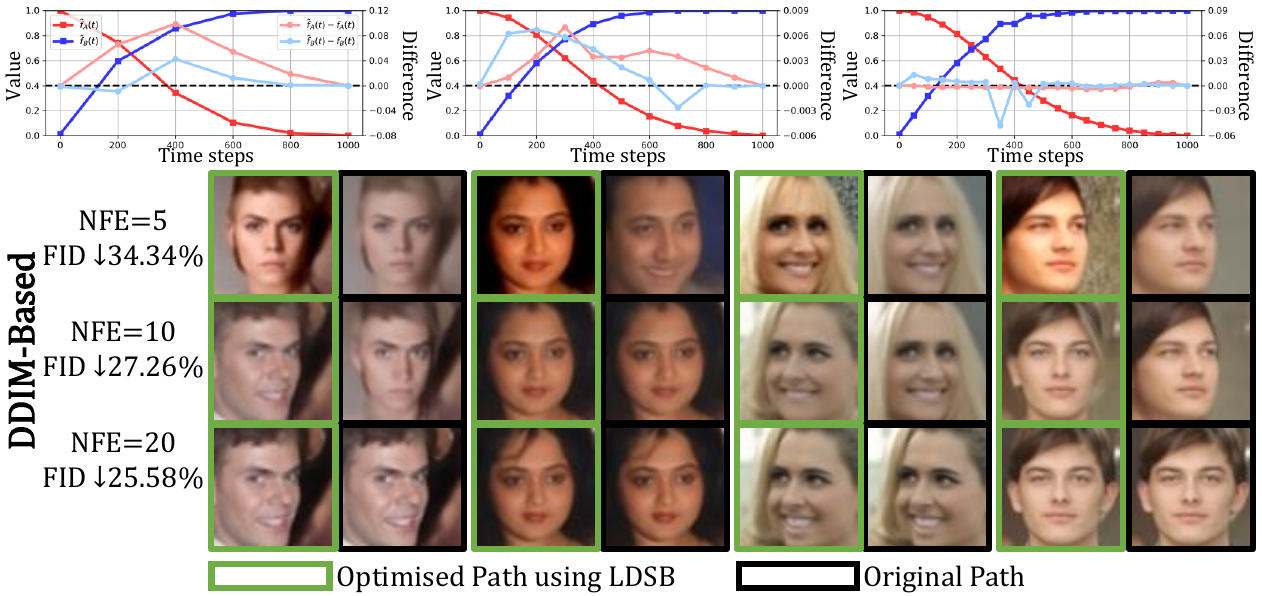}
	\caption{Comparison of diffusion paths and the generated image on CelebA. LDSB accelerates the image generation, as shown that images generated by DDIM transform along a fixed trajectory as the NFE increases, such as from front face to side face in the first column. Our LDSB accelerates this transformation.}
	\label{CelebA}
\end{figure}

\begin{figure}[t!]
	\centering
	\includegraphics[width=\columnwidth]{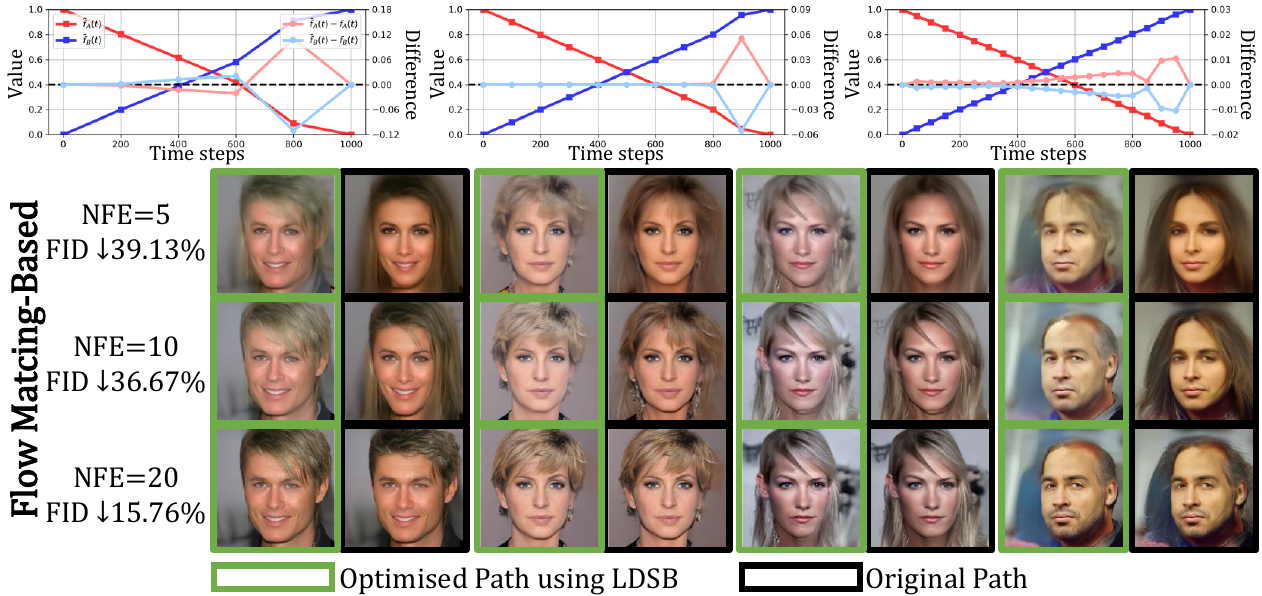}
	\caption{Comparison of the diffusion path and generated images on CelebA-HQ. LDSB accelerates the generation, as shown that the images generated by FM are always long-haired women with blurred background when NFE is $5$, whereas the LDSB optimised images have the correct gender and is more realistic.}
	\label{CelebA-HQ}
\end{figure}

LDSB significantly enhances the image generation quality by optimizing only the diffusion path while using the same denoising networks parameters. As shown in Tab. \ref{Tab:DDIM}, LDSB reduces the FID of DDIM by at least $25.58\%$ with the optimised path.  
In particular, it significantly reduces the FID by $48.50\%$ for 10-step sampling on CIFAR from $18.678$ to $9.619$, which is even lower than the original 20-step sampling result of $11.080$. For FM as shown in Tab. \ref{Tab:FM}, LDSB reduces the FID by at least $15.76\%$ and particularly $39.13\%$ for 5-step sampling CelebA-HQ. 
Toy experiments on 2D checkerboard show that LDSB enhances the FM and DDIM, and outperforms DSB as shown in Fig. \ref{checkerboard}.
Fig. \ref{Cifar10}, \ref{CelebA} and \ref{CelebA-HQ} intuitively illustrate the enhancement of LDSB on image generation quality and the optimization of diffusion paths. 
In CIFAR as shown in Fig. \ref{Cifar10}, the 5-step and 10-step images generated by DDIM and FM are blurred and lack real semantic information. 
However, the same denoising network with LDSB optimised diffusion path generates recognizable and semantically accurate images in $10$ or even $5$ steps with LDSB. Fig. \ref{CelebA} shows that LDSB on CelebA makes the colors in $5$ steps more realistic and vivid. 
In CelebA-HQ as shown in Fig. \ref{CelebA-HQ}, the images generated in $5$ steps share the similar features, such as blurred backgrounds and women with long brown hair.
In contrast after LDSB optimization, the images in $5$ steps are more realistic, with clearer backgrounds, and are closer to the images generated in $20$ steps. Thus, quantitative and qualitative comparisons both demonstrated that LDSB significantly improves image generation quality.

\begin{table}[tpb]
	\centering
	\caption{LDSB optimises diffusion paths based on DDIM significantly decreases FID metrics and improves image generation quality. LDSB with fewer NFE also achieve the original generation quality with more NFE, especially LDSB+DDIM on CIFAR10 and CelebA with $10$ steps has lower FID than the original DDIM with $20$ steps.}
	\label{Tab:DDIM}
	\begin{tabular}{c|c|cc} 
		\toprule
		\multicolumn{1}{c}{Dataset}&\multicolumn{1}{c}{NFE}& \multicolumn{1}{c}{Method} & \multicolumn{1}{c}{FID $\downarrow$}\\
		
		\midrule													
		\multirow{6}{*}{\rotatebox{90}{CIFAR10}}&\multirow{2}{*}{$5$}&DDIM& 44.394\\
		&&DDIM + LDSB& \textbf{26.929 (↓ 39.34\%)}\\
		&\multirow{2}{*}{$10$}&DDIM&18.678\\
		&&DDIM + LDSB& \textbf{9.619 (↓ 48.50\%)}\\
		&\multirow{2}{*}{$20$}&DDIM& 11.080\\
		&&DDIM + LDSB&  \textbf{7.073 (↓ 36.16\%)}\\
		\midrule
		\multirow{6}{*}{\rotatebox{90}{CelebA}}&\multirow{2}{*}{$5$}&DDIM&  28.568\\
		&&DDIM + LDSB& \textbf{18.757 (↓ 34.34\%)}\\
		&\multirow{2}{*}{$10$}&DDIM&18.042\\
		&&DDIM + LDSB&\textbf{13.123 (↓ 27.26\%)}\\
		&\multirow{2}{*}{$20$}&DDIM&13.419\\
		&&DDIM + LDSB&\textbf{9.987 (↓ 25.58\%)}\\
		\bottomrule
	\end{tabular}
	
\end{table}

\begin{table}[tpb]
	\centering
	\caption{LDSB greatly reduces the FID metric and improves image generation quality by optimising the diffusion path of flow matching (FM).}
	\label{Tab:FM}
	\begin{tabular}{c|c|cc} 
		\toprule
		\multicolumn{1}{c}{Dataset}&\multicolumn{1}{c}{NFE}& \multicolumn{1}{c}{Method} & \multicolumn{1}{c}{FID $\downarrow$}\\
		
		\midrule													
		\multirow{6}{*}{\rotatebox{90}{CIFAR10}}&\multirow{2}{*}{$5$}&FM&  35.493\\
		&&FM + LDSB&\textbf{ 25.380 (↓ 28.49\%)}\\
		&\multirow{2}{*}{$10$}&FM& 13.263\\
		&&FM + LDSB& \textbf{10.989 (↓ 17.15\%)}\\
		&\multirow{2}{*}{$20$}&FM&7.309\\
		&&FM + LDSB& \textbf{5.209 (↓ 28.73\%)}\\
		\midrule
		\multirow{6}{*}{\rotatebox{90}{CelebA-HQ}}&\multirow{2}{*}{$5$}&FM&115.177 \\
		&&FM + LDSB&\textbf{70.106 (↓ 39.13\%)} \\
		&\multirow{2}{*}{$10$}&FM&62.041 \\
		&&FM + LDSB&\textbf{39.292 (↓ 36.67\%)}\\
		&\multirow{2}{*}{$20$}&FM& 30.994\\
		&&FM + LDSB&\textbf{26.108 (↓ 15.76\%)}\\
		\bottomrule
	\end{tabular}
	
\end{table}

LDSB enhances the efficiency of image generation by optimizing only the diffusion path with the same denoising model parameters. 
Tab.\ref{Tab:DDIM} and ref{Tab:FM} show that after LDSB optimisation, the image generated in fewer steps is closer to or even exceeds the images generated in more steps along the original path.
Specifically for CIFAR and CelebA, the FID of 10-step DDIM+LDSB is lower than the 20-step DDIM.
The qualitative comparisons in Fig. \ref{Cifar10}, \ref{CelebA} and \ref{CelebA-HQ} further support this point.
In CIFAR (Fig. \ref{Cifar10}), it is obvious that LDSB enables both DDIM and FM to achieve higher image generation quality with fewer steps. In CelebA (Fig. \ref{CelebA}), DDIM generates images that undergo a fixed transformation as the NFE increases. For example, in the first column, DDIM takes $20$ steps to change a face from front to side.
After using LDSB, sampling is accelerated, achieving the same transformation in fewer steps, and the face changes from front to side in just $10$ steps. 
The rest examples in Fig. \ref{CelebA} demonstrate similar acceleration.
The acceleration for Flow Matching in CelebA-HQ is also evident as shown in Fig. \ref{CelebA-HQ}. 
When NFE is $5$, most images generated by FM share similar features, such as blurred background and long brown hair, which significantly differ from the real images. 
As NFE increases, the generated faces and backgrounds become more realistic. After LDSB optimization, fewer steps number yields more realistic images, especially the gender can be correctly identified  even with NFE of $5$, which is not possible with the original FM as shown in the 1st and 4th columns in Fig. \ref{CelebA-HQ}. In conclusion, both quantitative and qualitative comparisons  demonstrated that LDSB significantly accelerates image generation.

\section{Conclusion}
This paper is the first to optimise the diffusion paths of diffusion models in subspace based on the Schrödinger Bridge (SB) problem at a minimal training cost. Firstly, LDSB models the diffusion paths reparametrically to predict neighbouring time steps more accurately and robustly.
Secondly, LDSB proposes a new optimisation strategy for the weights within the diffusion paths based on IPF.
Quantitative and qualitative experiments demonstrate that our method significantly improves the image generation quality and efficiency. 

{
	\small
	\bibliographystyle{ieeenat_fullname}
	\bibliography{reference}

\begin{thebibliography}{31}
\providecommand{\natexlab}[1]{#1}
\providecommand{\url}[1]{\texttt{#1}}
\expandafter\ifx\csname urlstyle\endcsname\relax
  \providecommand{\doi}[1]{doi: #1}\else
  \providecommand{\doi}{doi: \begingroup \urlstyle{rm}\Url}\fi

\bibitem[Bansal et~al.(2024)Bansal, Borgnia, Chu, Li, Kazemi, Huang, Goldblum,
  Geiping, and Goldstein]{bansal2024cold}
Arpit Bansal, Eitan Borgnia, Hong-Min Chu, Jie Li, Hamid Kazemi, Furong Huang,
  Micah Goldblum, Jonas Geiping, and Tom Goldstein.
\newblock Cold diffusion: Inverting arbitrary image transforms without noise.
\newblock \emph{Advances in Neural Information Processing Systems}, 36, 2024.

\bibitem[Chen et~al.(2023)Chen, He, Zheng, Tan, and Zhu]{chen2023schrodinger}
Zehua Chen, Guande He, Kaiwen Zheng, Xu Tan, and Jun Zhu.
\newblock Schrodinger bridges beat diffusion models on text-to-speech
  synthesis.
\newblock \emph{arXiv preprint arXiv:2312.03491}, 2023.

\bibitem[Chetrite et~al.(2021)Chetrite, Muratore-Ginanneschi, and
  Schwieger]{chetrite2021schrodinger}
Rapha{\"e}l Chetrite, Paolo Muratore-Ginanneschi, and Kay Schwieger.
\newblock E. schr{\"o}dinger’s 1931 paper “on the reversal of the laws of
  nature”[“{\"u}ber die umkehrung der naturgesetze”, sitzungsberichte der
  preussischen akademie der wissenschaften, physikalisch-mathematische klasse,
  8 n9 144--153].
\newblock \emph{The European Physical Journal H}, 46:\penalty0 1--29, 2021.

\bibitem[De~Bortoli et~al.(2021)De~Bortoli, Thornton, Heng, and Doucet]{DSB}
Valentin De~Bortoli, James Thornton, Jeremy Heng, and Arnaud Doucet.
\newblock Diffusion schr{\"o}dinger bridge with applications to score-based
  generative modeling.
\newblock \emph{Advances in Neural Information Processing Systems},
  34:\penalty0 17695--17709, 2021.

\bibitem[Fortet(1940)]{fortet1940resolution}
Robert Fortet.
\newblock R{\'e}solution d'un syst{\`e}me d'{\'e}quations de m.
  schr{\"o}dinger.
\newblock \emph{Journal de Math{\'e}matiques Pures et Appliqu{\'e}es},
  19\penalty0 (1-4):\penalty0 83--105, 1940.

\bibitem[Gushchin et~al.(2024)Gushchin, Kholkin, Burnaev, and
  Korotin]{gushchin2024light}
Nikita Gushchin, Sergei Kholkin, Evgeny Burnaev, and Alexander Korotin.
\newblock Light and optimal schr{\"o}dinger bridge matching.
\newblock In \emph{Forty-first International Conference on Machine Learning},
  2024.

\bibitem[Heusel et~al.(2017)Heusel, Ramsauer, Unterthiner, Nessler, and
  Hochreiter]{heusel2017gans}
Martin Heusel, Hubert Ramsauer, Thomas Unterthiner, Bernhard Nessler, and Sepp
  Hochreiter.
\newblock Gans trained by a two time-scale update rule converge to a local nash
  equilibrium.
\newblock \emph{Advances in neural information processing systems}, 30, 2017.

\bibitem[Ho et~al.(2020)Ho, Jain, and Abbeel]{ho2020denoising}
Jonathan Ho, Ajay Jain, and Pieter Abbeel.
\newblock Denoising diffusion probabilistic models.
\newblock \emph{Advances in neural information processing systems},
  33:\penalty0 6840--6851, 2020.

\bibitem[Hoogeboom and Salimans(2022)]{hoogeboom2022blurring}
Emiel Hoogeboom and Tim Salimans.
\newblock Blurring diffusion models.
\newblock \emph{arXiv preprint arXiv:2209.05557}, 2022.

\bibitem[Huang(2024{\natexlab{a}})]{huang2024one}
Hanwen Huang.
\newblock One-step data-driven generative model via schr$\backslash$" odinger
  bridge.
\newblock \emph{arXiv preprint arXiv:2405.12453}, 2024{\natexlab{a}}.

\bibitem[Huang(2024{\natexlab{b}})]{huang2024schr}
Hanwen Huang.
\newblock Schr$\backslash$" odinger bridge based deep conditional generative
  learning.
\newblock \emph{arXiv preprint arXiv:2409.17294}, 2024{\natexlab{b}}.

\bibitem[Karras(2017)]{karras2017progressive}
Tero Karras.
\newblock Progressive growing of gans for improved quality, stability, and
  variation.
\newblock \emph{arXiv preprint arXiv:1710.10196}, 2017.

\bibitem[Karras et~al.(2022)Karras, Aittala, Aila, and
  Laine]{karras2022elucidating}
Tero Karras, Miika Aittala, Timo Aila, and Samuli Laine.
\newblock Elucidating the design space of diffusion-based generative models.
\newblock \emph{Advances in neural information processing systems},
  35:\penalty0 26565--26577, 2022.

\bibitem[Kolmogorov(1957)]{kolmogorov1957representation}
Andrei~Nikolaevich Kolmogorov.
\newblock On the representation of continuous functions of many variables by
  superposition of continuous functions of one variable and addition.
\newblock In \emph{Doklady Akademii Nauk}, pages 953--956. Russian Academy of
  Sciences, 1957.

\bibitem[Krizhevsky et~al.(2009)Krizhevsky, Hinton,
  et~al.]{krizhevsky2009learning}
Alex Krizhevsky, Geoffrey Hinton, et~al.
\newblock Learning multiple layers of features from tiny images.
\newblock 2009.

\bibitem[Kullback(1968)]{kullback1968probability}
Solomon Kullback.
\newblock Probability densities with given marginals.
\newblock \emph{The Annals of Mathematical Statistics}, 39\penalty0
  (4):\penalty0 1236--1243, 1968.

\bibitem[Lipman et~al.(2022)Lipman, Chen, Ben-Hamu, Nickel, and
  Le]{lipman2022flow}
Yaron Lipman, Ricky~TQ Chen, Heli Ben-Hamu, Maximilian Nickel, and Matt Le.
\newblock Flow matching for generative modeling.
\newblock \emph{arXiv preprint arXiv:2210.02747}, 2022.

\bibitem[Liu et~al.(2023)Liu, Vahdat, Huang, Theodorou, Nie, and
  Anandkumar]{liu2023i2sb}
Guan-Horng Liu, Arash Vahdat, De-An Huang, Evangelos~A Theodorou, Weili Nie,
  and Anima Anandkumar.
\newblock I2sb: image-to-image schr{\"o}dinger bridge.
\newblock In \emph{Proceedings of the 40th International Conference on Machine
  Learning}, pages 22042--22062, 2023.

\bibitem[Liu et~al.(2024{\natexlab{a}})Liu, Wang, Fan, Wang, Tang, and
  Qu]{liu2024residual}
Jiawei Liu, Qiang Wang, Huijie Fan, Yinong Wang, Yandong Tang, and Liangqiong
  Qu.
\newblock Residual denoising diffusion models.
\newblock In \emph{Proceedings of the IEEE/CVF Conference on Computer Vision
  and Pattern Recognition}, pages 2773--2783, 2024{\natexlab{a}}.

\bibitem[Liu et~al.(2022)Liu, Gong, and Liu]{liu2022flow}
Xingchao Liu, Chengyue Gong, and Qiang Liu.
\newblock Flow straight and fast: Learning to generate and transfer data with
  rectified flow.
\newblock \emph{arXiv preprint arXiv:2209.03003}, 2022.

\bibitem[Liu et~al.(2015)Liu, Luo, Wang, and Tang]{liu2015deep}
Ziwei Liu, Ping Luo, Xiaogang Wang, and Xiaoou Tang.
\newblock Deep learning face attributes in the wild.
\newblock In \emph{Proceedings of the IEEE international conference on computer
  vision}, pages 3730--3738, 2015.

\bibitem[Liu et~al.(2024{\natexlab{b}})Liu, Wang, Vaidya, Ruehle, Halverson,
  Solja{\v{c}}i{\'c}, Hou, and Tegmark]{liu2024kan}
Ziming Liu, Yixuan Wang, Sachin Vaidya, Fabian Ruehle, James Halverson, Marin
  Solja{\v{c}}i{\'c}, Thomas~Y Hou, and Max Tegmark.
\newblock Kan: Kolmogorov-arnold networks.
\newblock \emph{arXiv preprint arXiv:2404.19756}, 2024{\natexlab{b}}.

\bibitem[Luo et~al.(2023)Luo, Gustafsson, Zhao, Sj{\"o}lund, and
  Sch{\"o}n]{luo2023image}
Ziwei Luo, Fredrik~K Gustafsson, Zheng Zhao, Jens Sj{\"o}lund, and Thomas~B
  Sch{\"o}n.
\newblock Image restoration with mean-reverting stochastic differential
  equations.
\newblock In \emph{Proceedings of the 40th International Conference on Machine
  Learning}, pages 23045--23066, 2023.

\bibitem[Nichol and Dhariwal(2021)]{nichol2021improved}
Alexander~Quinn Nichol and Prafulla Dhariwal.
\newblock Improved denoising diffusion probabilistic models.
\newblock In \emph{International conference on machine learning}, pages
  8162--8171. PMLR, 2021.

\bibitem[R{\"u}schendorf and Thomsen(1993)]{ruschendorf1993note}
Ludger R{\"u}schendorf and Wolfgang Thomsen.
\newblock Note on the schr{\"o}dinger equation and i-projections.
\newblock \emph{Statistics \& probability letters}, 17\penalty0 (5):\penalty0
  369--375, 1993.

\bibitem[Schr{\"o}dinger(1932)]{schrodinger1932theorie}
Erwin Schr{\"o}dinger.
\newblock Sur la th{\'e}orie relativiste de l'{\'e}lectron et
  l'interpr{\'e}tation de la m{\'e}canique quantique.
\newblock In \emph{Annales de l'institut Henri Poincar{\'e}}, pages 269--310,
  1932.

\bibitem[Shi et~al.(2024)Shi, De~Bortoli, Campbell, and
  Doucet]{shi2024diffusion}
Yuyang Shi, Valentin De~Bortoli, Andrew Campbell, and Arnaud Doucet.
\newblock Diffusion schr{\"o}dinger bridge matching.
\newblock \emph{Advances in Neural Information Processing Systems}, 36, 2024.

\bibitem[Song et~al.(2020)Song, Meng, and Ermon]{song2020denoising}
Jiaming Song, Chenlin Meng, and Stefano Ermon.
\newblock Denoising diffusion implicit models.
\newblock \emph{arXiv preprint arXiv:2010.02502}, 2020.

\bibitem[Song and Ermon(2019)]{song2019generative}
Yang Song and Stefano Ermon.
\newblock Generative modeling by estimating gradients of the data distribution.
\newblock \emph{Advances in neural information processing systems}, 32, 2019.

\bibitem[Song et~al.()Song, Sohl-Dickstein, Kingma, Kumar, Ermon, and
  Poole]{songscore}
Yang Song, Jascha Sohl-Dickstein, Diederik~P Kingma, Abhishek Kumar, Stefano
  Ermon, and Ben Poole.
\newblock Score-based generative modeling through stochastic differential
  equations.
\newblock In \emph{International Conference on Learning Representations}.

\bibitem[Tang et~al.(2024)Tang, Hang, Gu, Chen, and Guo]{tang2024simplified}
Zhicong Tang, Tiankai Hang, Shuyang Gu, Dong Chen, and Baining Guo.
\newblock Simplified diffusion schr$\backslash$" odinger bridge.
\newblock \emph{arXiv preprint arXiv:2403.14623}, 2024.

\end{thebibliography}
}

\clearpage
\setcounter{page}{1}
\appendix
\section{Experimental setup}
This section presents the experimental setup for DDIM and FM applied to $x_\theta$ and $\epsilon_\theta$in the reverse path $\pi^{2n+2}$ as defined in Sec. 3.1. The LDSB models both the forward and reverse paths as follows:
\begin{equation}
	\begin{split}
		&\pi^{2n+1}: x_t^{2n+1} = \prod_{i=0}^t A^{2n+1}(i) x_{0_\theta} + \prod_{i=T}^t B^{2n+1}(i) \epsilon_\theta\\
		&\pi^{2n+2}: x_t^{2n+2} = \prod_{i=0}^t A^{2n+2}(i) x_{0} + \prod_{i=T}^t B^{2n+2}(i) \epsilon\\
		&\forall n: \pi^{2n+1},\pi^{2n+2} \in \mathcal{P}_{A,B}, t \in \{0,1,...,T\}
	\end{split}
\end{equation}
where the reverse path is start from $P_{prior}$ (Gaussian noise), the $x_{0_\theta}$ and $\epsilon_\theta$ are predicted by the denoising network $D_\theta$.

For DDIM, the output of the denoising network is predicted Gaussian noise, and $x_t$ is known. Hence, the $x_{0_\theta}$ and $\epsilon_\theta$ is :
\begin{equation}
	\begin{split}
		\epsilon_\theta&=D_\theta(x_t,t)\\
		x_{0_\theta}&=(x_t-\prod_{i=T}^t B(i) D_\theta(x_t,t))/\prod_{i=0}^t A(i)
	\end{split}
\end{equation}

For FM, the denoising network outputs the direction $x_0-x_T$, where the $x_0 \sim P_{data}$ and $x_T \sim P_{prior}$. The $x_t$ is known. Hence, the $x_{0_\theta}$ and $\epsilon_\theta$ is obtained after reparameterization :
\begin{equation}
	\begin{split}
		\epsilon_\theta&=\frac{x_t-\prod_{i=0}^t A(i) D_\theta(x_t,t)}{\prod_{i=0}^t A(i)+\prod_{i=T}^t B(i)}\\
		x_{0_\theta}&=\frac{x_t+\prod_{i=T}^t B(i) D_\theta(x_t,t)}{\prod_{i=0}^t A(i)+\prod_{i=T}^t B(i)}\\
	\end{split}
\end{equation}
Specifically, FM originally sets $f_A = t/T$ and $f_B = 1 - t/T$, as well as the sum of $f_A$ and $f_B$ is $1$. However, LDSB does not constrain the sum of $f_A$ and $f_B$ during prediction, which can directly affect the value range of $x_{0_\theta}$ and $\epsilon_\theta$ as denominators. Experiments found that sampling stability can be improved by scaling$f_A$ and $f_B$ isometrically to ensure their sum is $1$, as demonstrated in CelebA-HQ in FM.

\section{Results of Optimized Diffusion Paths}
The specific weight ($f_A$ and $f_B$) of the diffusion paths optimized by LDSB are provided in Tab. \ref{Tab:OTDDIM} and Tab. \ref{Tab:OTFM}.
The experiments utilize the pre-trained denoising network parameters officially provided by DDIM (CIFAR and CelebA) and Flow Matching (CIFAR and CelebA-HQ).
\begin{table*}[htbp]
	\centering
	\small
	\caption{Original diffusion path in DDIM and optimised diffusion path using LDSB}
	\label{Tab:OTDDIM}
	\begin{subtable}[t]{1\textwidth}
		\caption{5-Step path}
		\begin{tabularx}{\textwidth}{c|XX|XX|XX} 
			\toprule
			\multirow{2}{*}{Time Steps}&\multicolumn{2}{c}{DDIM (Origin)} & \multicolumn{2}{c}{CIFAR10 (LDSB)} & \multicolumn{2}{c}{CelebA (LDSB)}  \\
			
			&$f_{A}$&	$f_{B}$	&$f_{A}$&	$f_{B}$	&$f_{A}$&	$f_{B}$			\\
			\midrule		
			0	&	0.99995 	&	0.01000 	&	1.00000 	&	0.00923 	&	1.00000 	&	0.01142 	\\
			200	&	0.81015 	&	0.58622 	&	0.80274 	&	0.58399 	&	0.74435 	&	0.59543 	\\
			400	&	0.43997 	&	0.89801 	&	0.43314 	&	0.89487 	&	0.34136 	&	0.85574 	\\
			600	&	0.15990 	&	0.98713 	&	0.10833 	&	0.98532 	&	0.10544 	&	0.97506 	\\
			800	&	0.03883 	&	0.99925 	&	0.02153 	&	0.99570 	&	0.02064 	&	0.99842 	\\
			
			\bottomrule
		\end{tabularx}
	\end{subtable}
	\begin{subtable}[t]{1\textwidth}
		\caption{10-Step path}
		\begin{tabularx}{\textwidth}{c|XX|XX|XX} 
			\toprule
			\multirow{2}{*}{Time Steps}&\multicolumn{2}{c}{DDIM (Origin)} & \multicolumn{2}{c}{CIFAR10 (LDSB)} & \multicolumn{2}{c}{CelebA (LDSB)}  \\
			
			&$f_{A}$&	$f_{B}$	&$f_{A}$&	$f_{B}$	&$f_{A}$&	$f_{B}$		\\
			\midrule		
			0	&	0.99995 	&	0.01000 	&	1.00000 	&	0.01043 	&	1.00000 	&	0.00983 	\\
			100	&	0.94612 	&	0.32382 	&	0.94370 	&	0.32281 	&	0.94515 	&	0.31762 	\\
			200	&	0.81015 	&	0.58622 	&	0.80653 	&	0.58421 	&	0.80662 	&	0.57952 	\\
			300	&	0.62770 	&	0.77845 	&	0.62379 	&	0.77525 	&	0.62069 	&	0.77264 	\\
			400	&	0.43997 	&	0.89801 	&	0.43538 	&	0.89460 	&	0.43651 	&	0.89361 	\\
			500	&	0.27892 	&	0.96031 	&	0.27906 	&	0.95708 	&	0.27553 	&	0.95814 	\\
			600	&	0.15990 	&	0.98713 	&	0.16213 	&	0.97921 	&	0.15572 	&	0.98646 	\\
			700	&	0.08288 	&	0.99656 	&	0.07561 	&	0.98634 	&	0.07938 	&	0.99920 	\\
			800	&	0.03883 	&	0.99925 	&	0.02897 	&	0.98634 	&	0.03670 	&	0.99920 	\\
			900	&	0.01644 	&	0.99986 	&	0.00924 	&	1.00000 	&	0.01548 	&	1.00000 	\\
			
			\bottomrule
		\end{tabularx}
	\end{subtable}
	\begin{subtable}[t]{1\textwidth}
		\caption{20-Step path}
		\begin{tabularx}{\textwidth}{c|XX|XX|XX} 
			\toprule
			\multirow{2}{*}{Time Steps}&\multicolumn{2}{c}{DDIM (Origin)} & \multicolumn{2}{c}{CIFAR10 (LDSB)} & \multicolumn{2}{c}{CelebA (LDSB)}  \\
			
			&$f_{A}$&	$f_{B}$	&$f_{A}$&	$f_{B}$	&$f_{A}$&	$f_{B}$		\\
			\midrule		
			0	&	0.99995 	&	0.01000 	&	1.00000 	&	0.01018 	&	1.00000 	&	0.00928 	\\
			50	&	0.98486 	&	0.17335 	&	0.98477 	&	0.17398 	&	0.98500 	&	0.16057 	\\
			100	&	0.94612 	&	0.32382 	&	0.94510 	&	0.32440 	&	0.94755 	&	0.31583 	\\
			150	&	0.88650 	&	0.46272 	&	0.88500 	&	0.46257 	&	0.88846 	&	0.45574 	\\
			200	&	0.81015 	&	0.58622 	&	0.80802 	&	0.58566 	&	0.81150 	&	0.58104 	\\
			250	&	0.72210 	&	0.69179 	&	0.72006 	&	0.69098 	&	0.72364 	&	0.68790 	\\
			300	&	0.62770 	&	0.77845 	&	0.62547 	&	0.77755 	&	0.62969 	&	0.77391 	\\
			350	&	0.53215 	&	0.84665 	&	0.53080 	&	0.89658 	&	0.53403 	&	0.89456 	\\
			400	&	0.43997 	&	0.89801 	&	0.43963 	&	0.89658 	&	0.44200 	&	0.89456 	\\
			450	&	0.35474 	&	0.93497 	&	0.35278 	&	0.95804 	&	0.35683 	&	0.95803 	\\
			500	&	0.27892 	&	0.96031 	&	0.27576 	&	0.95804 	&	0.28106 	&	0.95803 	\\
			550	&	0.21386 	&	0.97686 	&	0.21368 	&	0.97286 	&	0.21615 	&	0.97421 	\\
			600	&	0.15990 	&	0.98713 	&	0.16443 	&	0.97447 	&	0.16307 	&	0.98435 	\\
			650	&	0.11658 	&	0.99318 	&	0.12495 	&	0.97539 	&	0.12089 	&	0.99406 	\\
			700	&	0.08288 	&	0.99656 	&	0.09488 	&	0.97583 	&	0.08692 	&	0.99729 	\\
			750	&	0.05745 	&	0.99835 	&	0.07023 	&	0.97625 	&	0.06112 	&	0.99787 	\\
			800	&	0.03883 	&	0.99925 	&	0.04959 	&	0.97625 	&	0.04030 	&	0.99787 	\\
			850	&	0.02559 	&	0.99967 	&	0.03349 	&	0.97668 	&	0.02365 	&	0.99799 	\\
			900	&	0.01644 	&	0.99986 	&	0.02109 	&	0.97761 	&	0.01282 	&	0.99881 	\\
			950	&	0.01030 	&	0.99995 	&	0.01228 	&	1.00000 	&	0.00682 	&	1.00000 	\\

			\bottomrule
		\end{tabularx}
	\end{subtable}
	
\end{table*}

\begin{table*}[htpb]
	\centering
	\caption{Original diffusion path in Flow Matching and optimised diffusion path using LDSB}
	\small
	\label{Tab:OTFM}
	\begin{subtable}[t]{1\textwidth}
		\caption{5-Step path}
		\begin{tabularx}{\textwidth}{c|XX|XX|XX} 
			\toprule
			\multirow{2}{*}{Time Steps}&\multicolumn{2}{c}{Flow Matching (Origin)} & \multicolumn{2}{c}{CIFAR10 (LDSB)} & \multicolumn{2}{c}{CelebA (LDSB)}  \\
			
			&$f_{A}$&	$f_{B}$	&$f_{A}$&	$f_{B}$	&$f_{A}$&	$f_{B}$		\\
			\midrule		
			0	&	1.00000 	&	0.00000 	&	1.00000 	&	0.00000 	&	1.00000 	&	0.00000 	\\
			200	&	0.80000 	&	0.20000 	&	0.77570 	&	0.20977 	&	0.80200 	&	0.19800 	\\
			400	&	0.60000 	&	0.40000 	&	0.58961 	&	0.40658 	&	0.61247 	&	0.38753 	\\
			600	&	0.40000 	&	0.60000 	&	0.40095 	&	0.60201 	&	0.42058 	&	0.57942 	\\
			800	&	0.20000 	&	0.80000 	&	0.05246 	&	0.80053 	&	0.08983 	&	0.91017 	\\

			\bottomrule
		\end{tabularx}
	\end{subtable}
	\begin{subtable}[t]{1\textwidth}
		\caption{10-Step path}
		\begin{tabularx}{\textwidth}{c|XX|XX|XX} 
			\toprule
			\multirow{2}{*}{Time Steps}&\multicolumn{2}{c}{Flow Matching (Origin)} & \multicolumn{2}{c}{CIFAR10 (LDSB)} & \multicolumn{2}{c}{CelebA (LDSB)}\\
			
			&$f_{A}$&	$f_{B}$	&$f_{A}$&	$f_{B}$	&$f_{A}$&	$f_{B}$		\\
			\midrule		
			0	&	1.00000 	&	0.00000 	&	1.00000 	&	0.00000 	&	1.00000 	&	0.00000 	\\
			100	&	0.90000 	&	0.10000 	&	0.88914 	&	0.10762 	&	0.90011 	&	0.09989 	\\
			200	&	0.80000 	&	0.20000 	&	0.78573 	&	0.20539 	&	0.79990 	&	0.20010 	\\
			300	&	0.70000 	&	0.30000 	&	0.67939 	&	0.30450 	&	0.69992 	&	0.30008 	\\
			400	&	0.60000 	&	0.40000 	&	0.57500 	&	0.40346 	&	0.60003 	&	0.39997 	\\
			500	&	0.50000 	&	0.50000 	&	0.47184 	&	0.50230 	&	0.49999 	&	0.50001 	\\
			600	&	0.40000 	&	0.60000 	&	0.36996 	&	0.60141 	&	0.40011 	&	0.59989 	\\
			700	&	0.30000 	&	0.70000 	&	0.26685 	&	0.70045 	&	0.29987 	&	0.70013 	\\
			800	&	0.20000 	&	0.80000 	&	0.15307 	&	0.80057 	&	0.19913 	&	0.80087 	\\
			900	&	0.10000 	&	0.90000 	&	0.05306 	&	0.89695 	&	0.04486 	&	0.95514 	\\

			\bottomrule
		\end{tabularx}
	\end{subtable}
	\begin{subtable}[t]{1\textwidth}
		\caption{20-Step path}
		\begin{tabularx}{\textwidth}{c|XX|XX|XX} 
			\toprule
			\multirow{2}{*}{Time Steps}&\multicolumn{2}{c}{Flow Matching (Origin)} & \multicolumn{2}{c}{CIFAR10 (LDSB)} & \multicolumn{2}{c}{CelebA (LDSB)}  \\
			
			&$f_{A}$&	$f_{B}$	&$f_{A}$&	$f_{B}$	&$f_{A}$&	$f_{B}$			\\
			\midrule		
			0	&	1.00000 	&	0.00000 	&	1.00000 	&	0.00000 	&	1.00000 	&	0.00000 	\\
			50	&	0.95000 	&	0.05000 	&	0.94195 	&	0.06487 	&	0.94874 	&	0.05126 	\\
			100	&	0.90000 	&	0.10000 	&	0.87925 	&	0.11170 	&	0.89906 	&	0.10094 	\\
			150	&	0.85000 	&	0.15000 	&	0.82802 	&	0.15753 	&	0.84911 	&	0.15089 	\\
			200	&	0.80000 	&	0.20000 	&	0.77480 	&	0.21011 	&	0.79919 	&	0.20081 	\\
			250	&	0.75000 	&	0.25000 	&	0.71373 	&	0.25986 	&	0.74926 	&	0.25074 	\\
			300	&	0.70000 	&	0.30000 	&	0.66295 	&	0.30991 	&	0.69949 	&	0.30051 	\\
			350	&	0.65000 	&	0.35000 	&	0.60922 	&	0.36024 	&	0.64945 	&	0.35055 	\\
			400	&	0.60000 	&	0.40000 	&	0.55540 	&	0.41058 	&	0.59907 	&	0.40093 	\\
			450	&	0.55000 	&	0.45000 	&	0.49805 	&	0.45857 	&	0.54852 	&	0.45148 	\\
			500	&	0.50000 	&	0.50000 	&	0.44299 	&	0.50801 	&	0.49825 	&	0.50175 	\\
			550	&	0.45000 	&	0.55000 	&	0.38999 	&	0.55719 	&	0.44753 	&	0.55247 	\\
			600	&	0.40000 	&	0.60000 	&	0.33685 	&	0.60583 	&	0.39704 	&	0.60296 	\\
			650	&	0.35000 	&	0.65000 	&	0.28416 	&	0.65499 	&	0.34660 	&	0.65340 	\\
			700	&	0.30000 	&	0.70000 	&	0.23635 	&	0.70432 	&	0.29606 	&	0.70394 	\\
			750	&	0.25000 	&	0.75000 	&	0.17714 	&	0.75340 	&	0.24553 	&	0.75447 	\\
			800	&	0.20000 	&	0.80000 	&	0.12667 	&	0.80305 	&	0.19562 	&	0.80438 	\\
			850	&	0.15000 	&	0.85000 	&	0.08734 	&	0.85257 	&	0.14861 	&	0.85139 	\\
			900	&	0.10000 	&	0.90000 	&	0.03767 	&	0.90136 	&	0.09043 	&	0.90957 	\\
			950	&	0.05000 	&	0.95000 	&	0.00316 	&	0.95189 	&	0.03957 	&	0.96043 	\\

			\bottomrule
		\end{tabularx}
	\end{subtable}
	
\end{table*}

\begin{figure*}[t]
	\centering
	\includegraphics[width=.93\textwidth]{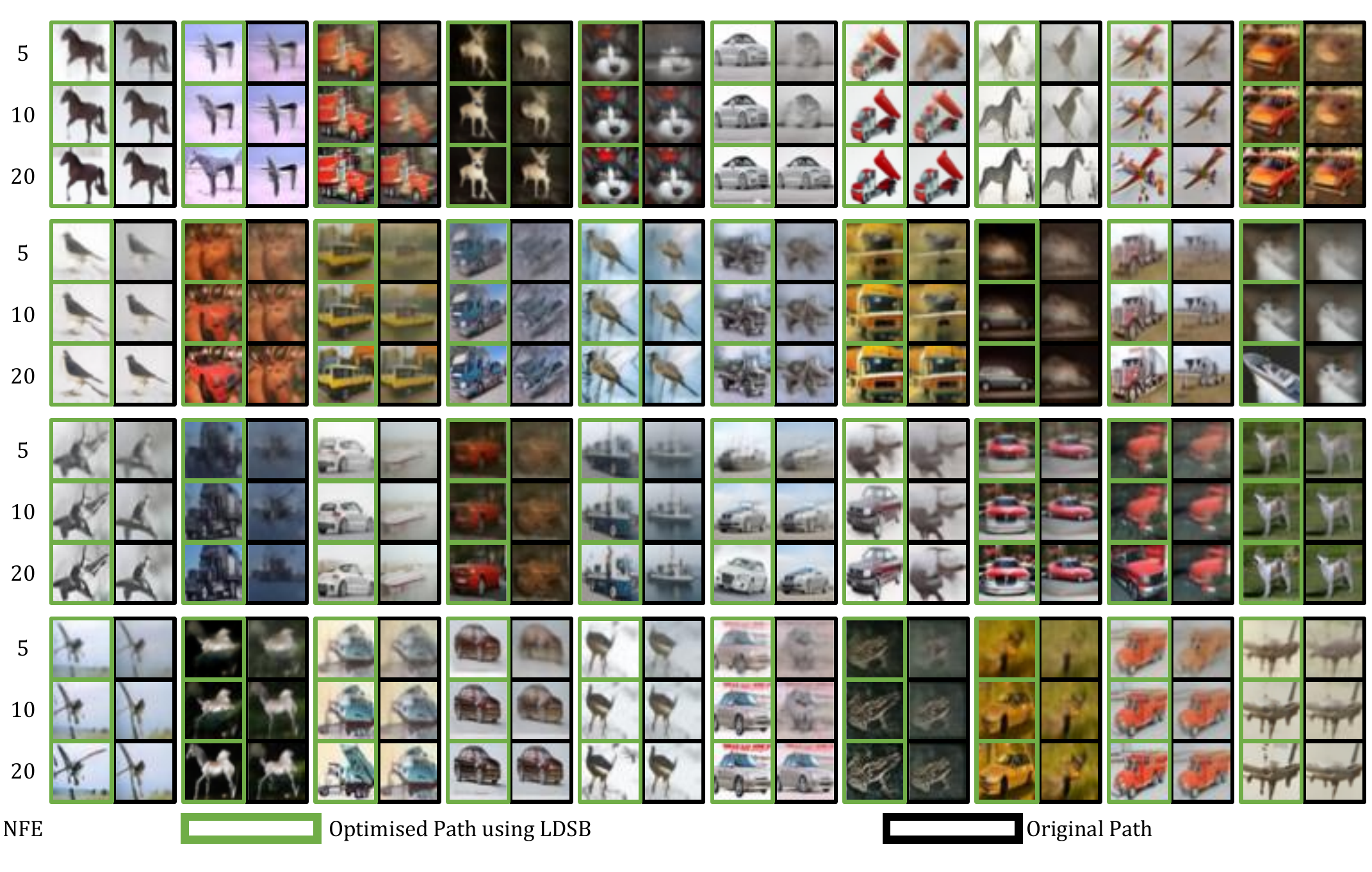}
	\caption{Comparison results of LDSB optimisation over DDIM on CIFAR10 $32\times32$. The NFE is 5, 10 and 20.}
	\label{CifarDDIM1}
\end{figure*}
\begin{figure*}[htbp]
	\centering
	\includegraphics[width=.93\textwidth]{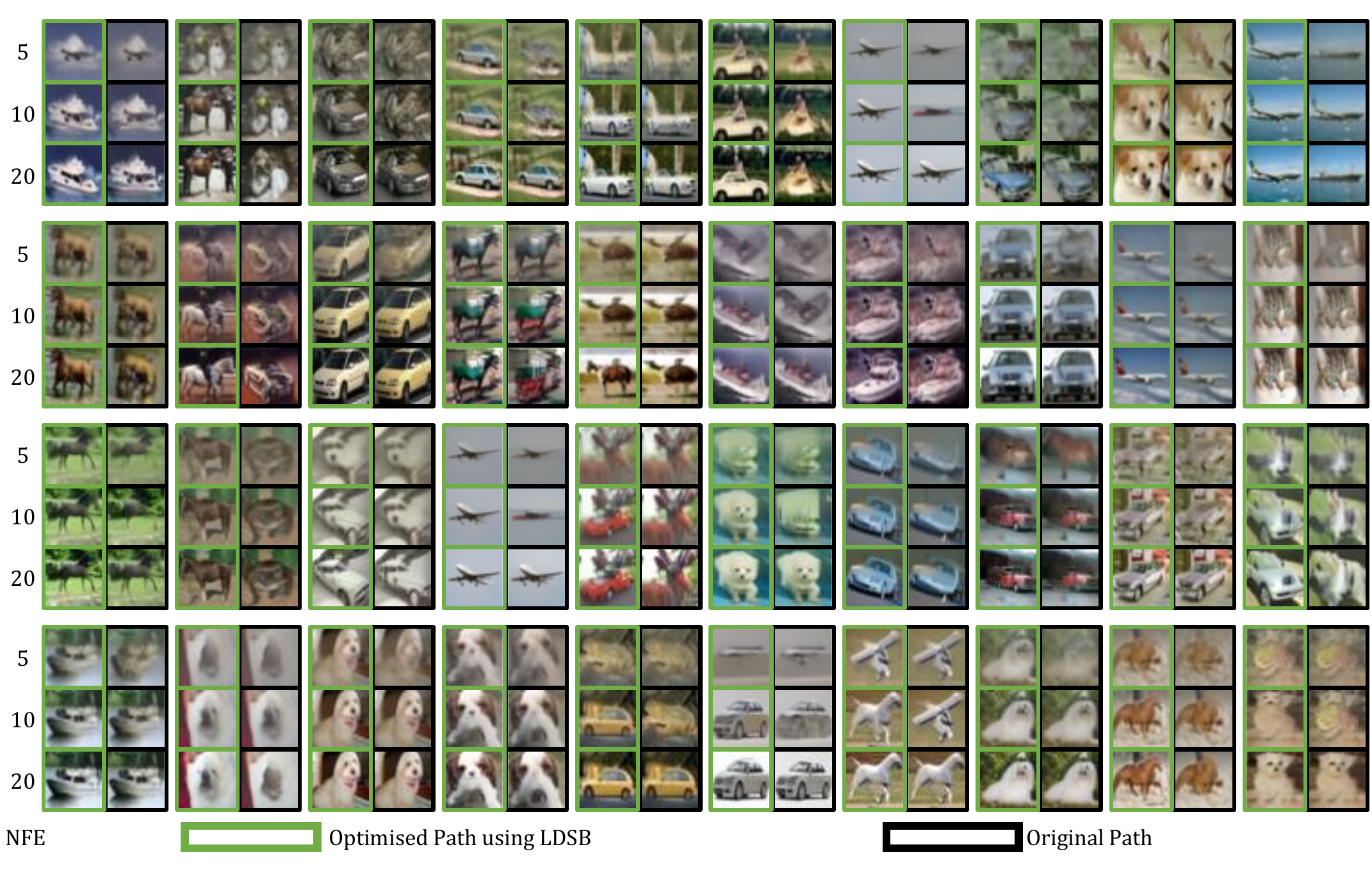}
	\caption{Comparison results of LDSB optimisation over Flow Matching on CIFAR10 $32\times32$. The NFE is 5, 10 and 20.}
	\label{CifarFM1}
\end{figure*}
\begin{figure*}[htbp]
	\centering
	\includegraphics[width=\textwidth]{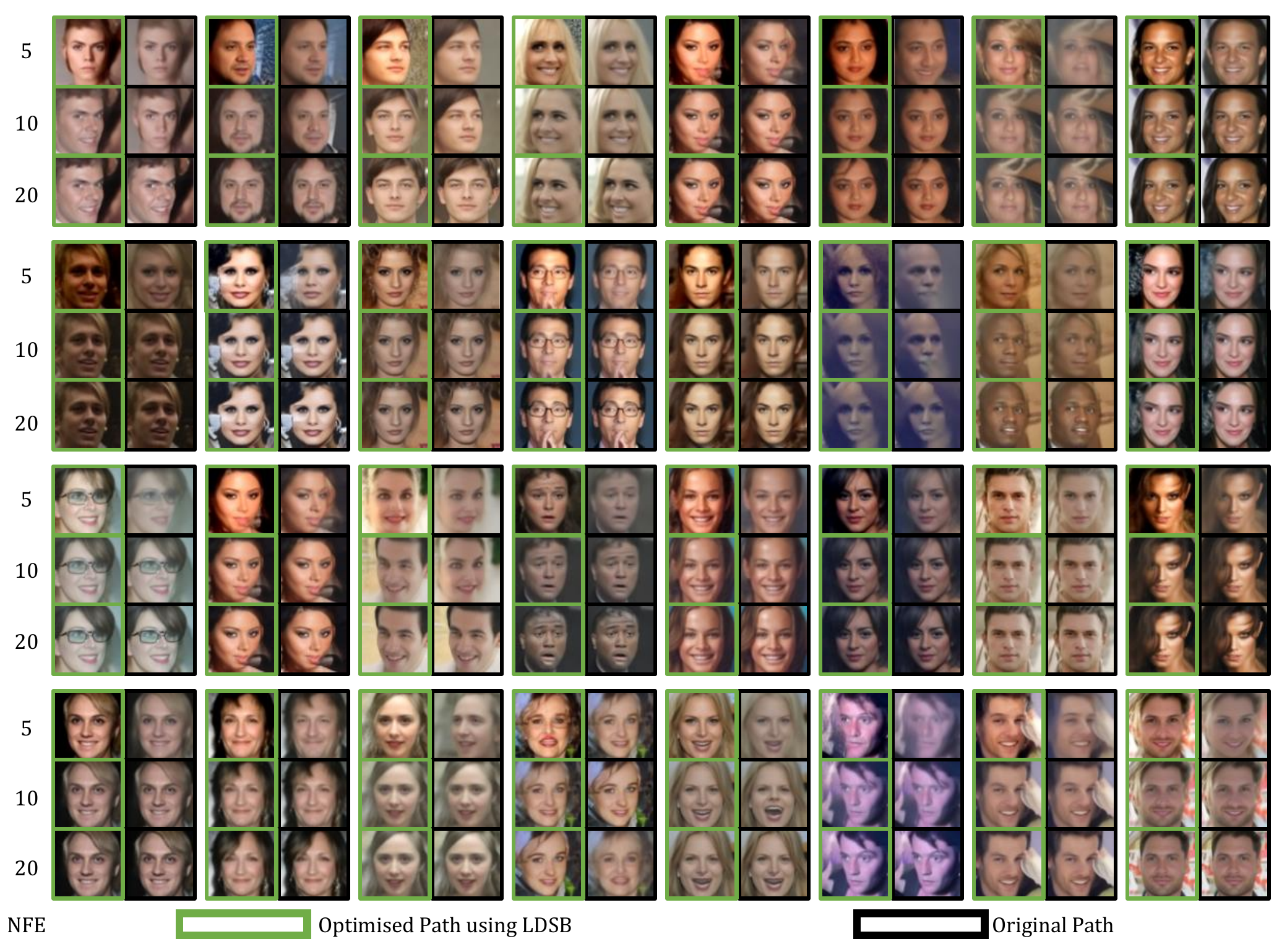}
	\caption{Comparison results of LDSB optimisation over DDIM on CelebA $64\times64$. The NFE is 5, 10 and 20.}
	\label{CelebA1}
\end{figure*}
\begin{figure*}[htbp]
	\centering
	\includegraphics[width=\textwidth]{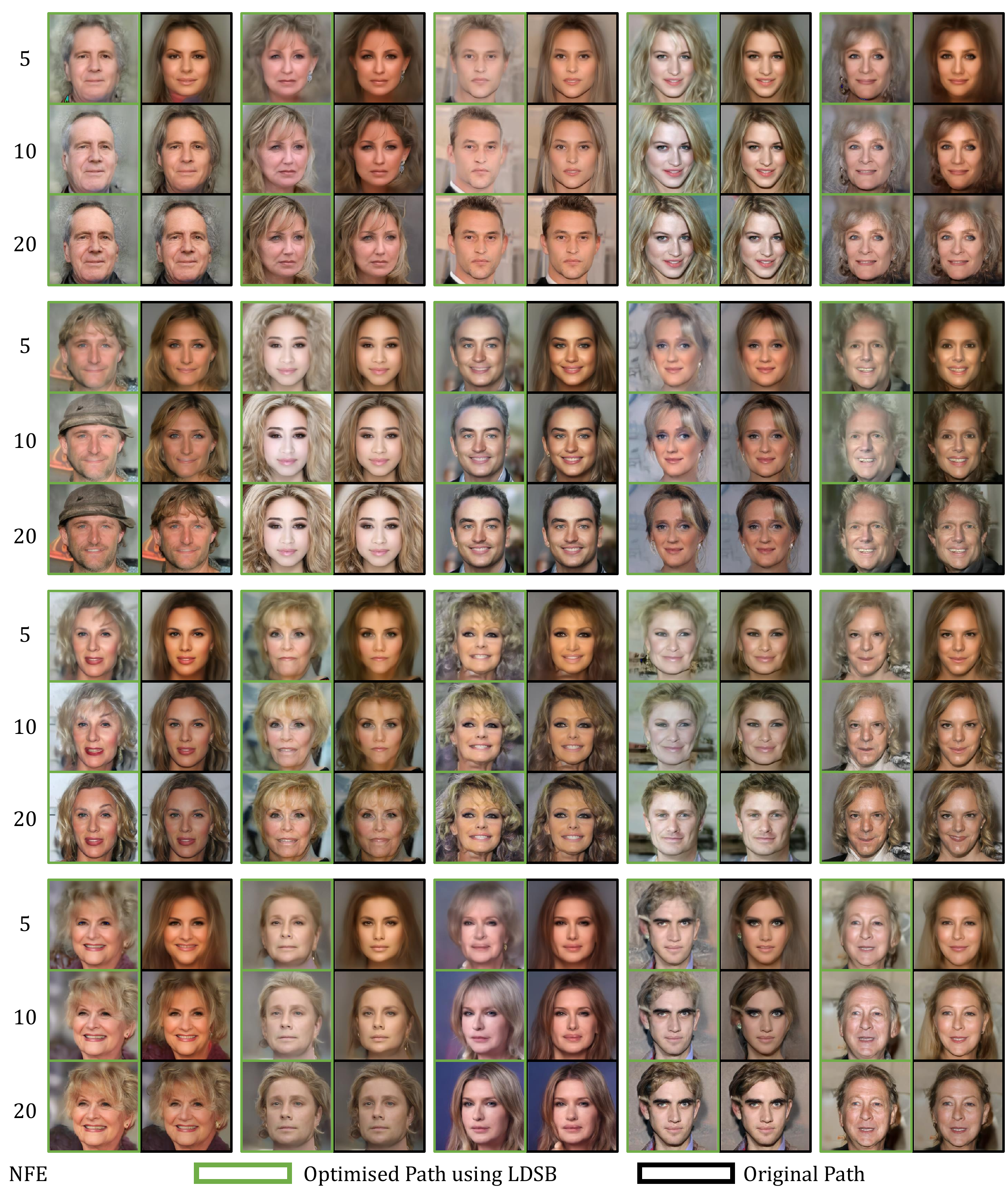}
	\caption{Comparison results of LDSB optimisation over Flow Matching on CelebA-HQ $256\times256$. The NFE is 5, 10 and 20.}
	\label{CelebA-HQ1}
\end{figure*}

\end{document}